%% file: main.tex
\documentclass{article}

\input{control.tex}

\makeatletter
\let\blx@rerun@biber\relax
\makeatother

\input{header.tex}

\begin{document}
\maketitle
\input{acronyms.tex}

\input{sections/00_abstract.tex}
\input{sections/01_introduction.tex}
\input{sections/03_model.tex}

\input{sections/04_inference.tex}

\input{sections/05_experiments.tex}
\input{sections/06_conclusion.tex}

\input{sections/07_acknowledgments.tex}

\small
\newrefcontext[labelprefix=]
\printbibliography[keyword=main]
\normalsize

\appendix
\newpage
\input{appendix/A0_header.tex}
\input{appendix/A1_appendix.tex}

\small
\newrefcontext[labelprefix=A]
\printbibliography[keyword=appendix, resetnumbers=true]

\end{document}

%% file: control.tex
\usepackage[nonatbib,preprint]{neurips_2021} 

\makeatletter
\newcommand{\maketitlenew}{\@maketitle}
\makeatother

\usepackage[utf8]{inputenc} 
\usepackage[T1]{fontenc}    
\usepackage{hyperref}       
\usepackage{url}            
\usepackage{booktabs}       
\usepackage{amsfonts}       
\usepackage{nicefrac}       
\usepackage{microtype}      
\usepackage{xcolor}         
\usepackage{graphicx}
\usepackage{subfigure}

\usepackage[backend=biber, sorting=none, defernumbers=true, maxbibnames=99]{biblatex}
\DeclareSourcemap{
  \maps[datatype=bibtex, overwrite]{
    \map{
    \perdatasource{main.bib}
     \step[fieldset=KEYWORDS, fieldvalue=main, append]
   }
   \map{
    \perdatasource{appendix.bib}
     \step[fieldset=KEYWORDS, fieldvalue=appendix, append]
   }
 }
}

\usepackage{amssymb}
\usepackage{amsmath}
\usepackage{esint}           
\usepackage{bbm}

\usepackage{wrapfig}
\usepackage[plainruled,titlenumbered]{algorithm2e}
\makeatletter
\newcommand{\removelatexerror}{\let\@latex@error\@gobble}
\makeatother
\usepackage{listings}
\usepackage{tikz}
\usetikzlibrary{bayesnet}
\usetikzlibrary{positioning}
\usetikzlibrary{arrows.meta}
\usetikzlibrary{calc}

\usetikzlibrary{positioning,fit,shapes,backgrounds,circuits.logic.US}

\usepackage[shortlabels,inline]{enumitem}  
\newlist{inlineitemize}{enumerate*}{1}
\setlist[inlineitemize]{label=(\roman*)}

\usepackage[capitalise]{cleveref} 

\usepackage{todonotes}
\newboolean{todo}
\setboolean{todo}{true} 

\newcommand{\remarkInternal}[4]{\ifthenelse{\boolean{todo}}{\todo[inline, color=#2, caption={2do}, #3]{\begin{minipage}{\textwidth-4pt}\emph{Remark #1:}\\#4\end{minipage}}}{}}

\DeclareMathOperator{\tr}{\mathrm{tr}\,}

\DeclareMathOperator*{\argmax}{\arg\max}

\DeclareMathOperator{\1}{\mathbbm{1} }

\DeclareMathOperator{\E}{\mathsf{E}\,}

\DeclareMathOperator{\KL}{\mathrm{KL}\,}

\DeclareMathOperator{\NDis}{\mathcal{N}\,}

\usepackage[nolist]{acronym} 

%% file: header.tex
\title{Variational Inference for Continuous-Time\\ Switching Dynamical Systems}
\author{\begin{NoHyper}Lukas~Köhs \qquad Bastian~Alt \qquad Heinz~Koeppl\end{NoHyper} \\
  Department of Electrical Engineering and Information Technology\\
  Technische Universität Darmstadt\\
\texttt{\{lukas.koehs, bastian.alt,  heinz.koeppl\}@bcs.tu-darmstadt.de} 
}

%% file: acronyms.tex
\begin{acronym}
\acro{ssa}[SSA]{stochastic sampling algorithm}
\acro{ctmc}[CTMC]{continuous-time Markov chain}
\acroplural{ctmc}[CTMCs]{continuous-time Markov chains}
\acro{sde}[SDE]{stochastic differential equation}
\acroplural{sde}[SDEs]{stochastic differential equations}
\acro{em}[EM]{expectation maximization}
\acro{vem}[VEM]{variational expectation maximization}
\acro{slds}[SLDS]{switching linear dynamical system}
\acroplural{slds}[SLDS]{switching linear dynamical system}
\acro{pde}[PDE]{partial differential equation}
\acroplural{pde}[PDEs]{partial differential equations}
\acro{ode}[ODE]{ordinary differential equation} 
\acroplural{ode} [ODEs]{ordinary differential equations} 
\acro{sde}[SDE]{stochastic differential equation} 
\acroplural{sde} [SDEs]{stochastic differential equations}
\acro{ssde}[SSDE]{switching stochastic differential equation} 
\acroplural{ssde}[SSDEs]{switching stochastic differential equations}
\acro{mjp}[MJP]{Markov jump process}
\acroplural{mjp}[MJPs]{Markov jump processes}
\acro{gp}[GP]{Gaussian process}
\acroplural{gp}[GPs]{Gaussian processes}
\acro{gpa}[GPA]{Gaussian process approximation}
\acro{kl}[KL]{Kullback-Leibler}
\acro{fpe}[FPE]{Fokker-Planck equation}
\acro{gfpe}[HME]{hybrid master equation}
\acro{md}[MD]{molecular dynamics}
\acro{hmm}[HMM]{hidden Markov model}
\acroplural{hmm}[HMMs]{hidden Markov models}
\acro{pdf}[PDF]{probability density function}
\acro{cdf}[CDF]{cumulative distribution function}
\acro{msm}[MSM]{Markov state model}
\acroplural{msm}[MSMs]{Markov state models}
\acro{vi}[VI]{variational inference}
\acro{map}[MAP]{maximum a-posteriori}
\acro{hme}[HME]{hybrid master equation}
\acro{el}[EL]{Euler-Lagrange}
\acro{elbo}[ELBO]{evidence lower bound}
\end{acronym}

%% file: sections/00_abstract.tex
\begin{abstract}
Switching dynamical systems provide a powerful, interpretable modeling  framework for
 inference in time-series data in, e.g., the natural sciences or engineering applications.
Since many areas, such as biology or discrete-event systems, are naturally described in
 continuous time, we present a model based on an \acl{mjp} modulating a subordinated diffusion process. 
We provide the exact evolution equations for the prior and posterior marginal densities, 
 the direct solutions of which are however computationally intractable.
Therefore, we develop a new continuous-time variational inference algorithm, combining a 
 \acl{gp} approximation on the diffusion level with posterior inference for \aclp{mjp}. 
By minimizing the path-wise \acl{kl} divergence we obtain
\begin{inlineitemize}
\item Bayesian latent state estimates for arbitrary points on the real axis and
\item point estimates of unknown system parameters, utilizing \acl{vem}.
\end{inlineitemize}
We extensively evaluate our algorithm under the model assumption and for real-world examples.
\end{abstract}

%% file: sections/01_introduction.tex
\section{Introduction}
\label{introduction}
Many natural and engineered dynamical systems can be understood in terms of
continuous-discrete hybrid models, in which a given system switches between discrete modes
exhibiting continuous dynamics.
Examples include neuro-mechanical models of locomotion in neuroscience \cite{proctor2010},
state-dependent volatility dynamics in financial analysis
\cite{carvalhoSimulationbasedSequentialAnalysis2007} or electric power systems \cite{willsky1976},
and phenotype differentiation in systems biology \cite{bressloffStochasticSwitchingBiology2017}.

In a discrete-time setting, a widely used class of stochastic hybrid models are \ac{slds} 
\cite{barber2012}, which have received considerable attention
in recent years \cite{willskyNonparametricBayesianLearning2009,NIPS2016_6379,lindermanBayesianLearningInference2017,glaser2020}. 
Since real-world physical and biological systems naturally evolve in continuous time,
a discrete-time description of such systems is however limiting. In biological experiments and
discrete-event systems in engineering \cite{cassandras2009introduction},
for instance, one typically \begin{inlineitemize} \item is interested in the system behavior at 
any given point in time and \item can not easily determine an appropriate time discretization 
\end{inlineitemize}. 
To overcome these limitations, a continuous-time analog to \ac{slds} models has been put
forward termed \acp{ssde} 
\cite{xuerongStochasticDifferentialEquations2006}, which augment a set of diffusion
processes with an underlying \ac{mjp}.
Hybrid models of this kind have a long tradition in statistics \cite{davis2018markov}
and have been analyzed in particular in applications to biological systems
\cite{bressloff2020switching}.

Diffusion processes and \acp{mjp} have been treated extensively
in the literature \cite{ethierMarkovProcessesCharacterization2005,oksendal2003stochastic}.
For each process class individually, exact expressions for the posterior paths given 
some set of observations can be obtained \cite{andersonSmoothingAlgorithmsNonlinear1983}.
However, for diffusion processes in particular, these expressions quickly become 
intractable as they entail solving multi-dimensional \acp{pde}.
This is aggravated if the diffusion is coupled to an underlying jump process,
because both processes then have to be solved jointly.

One way to circumvent this issue are Monte Carlo approaches, which
have been devised for both diffusion \cite{ruiz2017particle} and jump processes
\cite{rao2013fast}. Sampling methods may however suffer from slow convergence, and,
as we shall show, still face computational intractabilities for hybrid systems. 
An established approach avoiding these problems
are \ac{vi} methods, which approximate the exact posteriors by optimization \cite{bleiVariationalInferenceReview2017}.
For inference and parameter learning in diffusion processes,
continuous-time \ac{vi} frameworks have been developed utilizing, e.g., \acp{gp} \cite{archambeauGaussianProcessApproximations2007,archambeauVariationalInferenceDiffusion2008},
and general exponential family distributions \cite{sutter2016}.
Similar methods have also been devised for inference in \acp{mjp}
\cite{opperVariationalInferenceMarkov2008,wildnerMomentBasedVariationalInference2019}.
To the best of our knowledge, an inference framework for continuous-time hybrid processes
is however lacking. We draw on these previous works and present a generalized
\ac{vi} framework for hybrid systems which recovers existing diffusion and \ac{mjp}
approximations as special cases. We specifically focus on meta-stable systems which
remain in distinct, qualitatively different regimes over extended periods of time,
which are of special interest, e.g., in computational structural biology \cite{weinan2004metastability}.
An implementation of our proposed method is publicly available\footnote{A link to the implementation will be inserted here in the camera-ready version of the paper.}.




%% file: sections/03_model.tex
\section{Mathematical Background}
\label{sec:model}
\subsection{The Model}
\begin{figure}
    \includegraphics[width=\columnwidth]{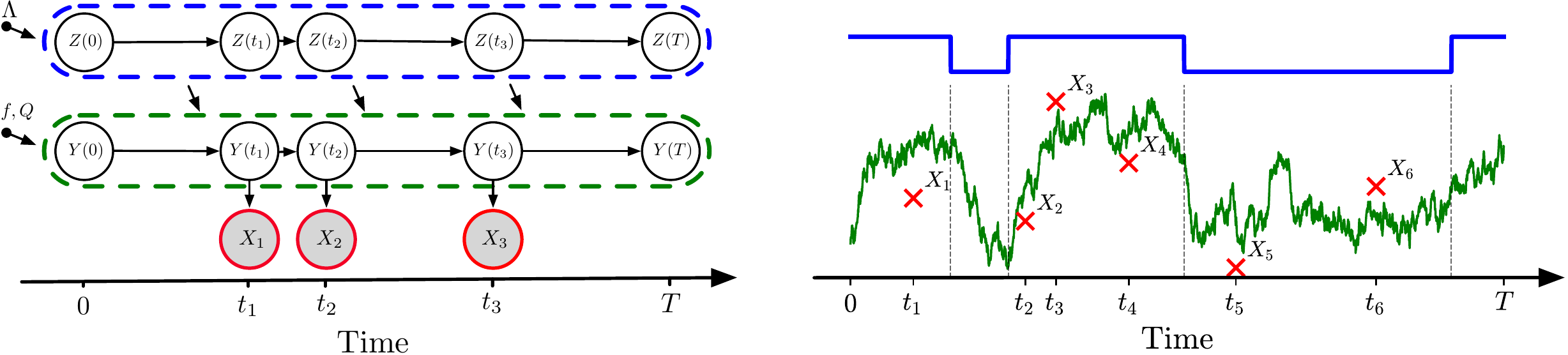}
    \caption{Sketch of the three layers of the hybrid process model. 
    A two-state \acs{mjp}  $Z(t)$ (blue) evolves
    freely in the time interval $t\in[0, T]$, see \cref{eq_transition_rates}
    .
    The \acs{mjp} controls the dynamics
    of the \ac{ssde} $Y(t)\mid Z(t)$ (green),  see \cref{eq:SSDE}.
    From these continuous dynamics, only noisy observations $X_1,X_2,\dots$ (red) are available for inference at
    irregularly-spaced time points $t_1, t_2, \dots$.
    Left: Graphical model.
    Right: Sample path (vertical dashed lines indicate the
    $Z$-transitions).
}
    \label{fig:process_sketch_pgm}
\end{figure}

In this work, we consider three joint stochastic processes $\{Z(t)\}_{t\geq 0}$,
$\{Y(t)\}_{t \geq 0}$, and $\{X_i\}_{i \in \mathbb N}$. A continuous-time adaptation of a 
probabilistic graphical model and a realization of the processes is shown in 
\cref{fig:process_sketch_pgm}. 
\paragraph{The Switching Process.}
The discrete-valued process $Z(t) \in \mathcal{Z} \subseteq \mathbb N$ is given as a latent \acf{mjp} 
freely evolving in time $t$. An \ac{mjp} is a continuous-time Markov process 
\cite{norrisMarkovChains1997} on a countable state space $\mathcal{Z}$ and
is completely characterized by an initial probability
$p_0(z) := \mathbb P(Z(0)=z), \; \forall z \in \mathcal Z$ and a transition rate function
\begin{equation}
    \Lambda(z, z',t) = \lim_{h\searrow 0}\frac{\mathbb P (Z(t+h)=z'\mid Z(t)=z)}{h}
    \label{eq_transition_rates}
\end{equation}
for $z' \in \mathcal Z \setminus z$, with the \emph{exit rate} $\Lambda(z,t):= \sum_{z'\in \mathcal Z \setminus z}\Lambda(z, z',t)$. 

\paragraph{The Subordinated Diffusion Process.}
The freely evolving \ac{mjp} controls the dynamics of a continuous-valued process
$Y(t) \in \mathcal Y \subseteq \mathbb R^n$, which is given as a latent \acf{ssde}
in an It\^{o} sense,
\cite{xuerongStochasticDifferentialEquations2006}
\begin{equation}
    \mathrm d Y(t) = f(Y(t), Z(t), t) \,\mathrm d t + Q(Y(t),Z(t), t) \,\mathrm d W(t),
    \label{eq:SSDE}
\end{equation}
where $f: \mathcal Y \times \mathcal Z \times \mathbb R_{\geq 0} \rightarrow \mathcal Y$ 
is an arbitrary drift function, 
the dispersion $Q: \mathcal{Y} \times \mathcal Z \times \mathbb R_{\geq 0} \rightarrow \mathbb R^{n \times m}$ 
determines the noise characteristics of the system and $W(t) \in \mathbb R^m$ is a standard Brownian motion.
We define the system noise covariance $D(Y(t), Z(t), t)$ as $D := QQ^\top$.
Note that the only difference to a conventional \ac{sde} is the $Z(t)$-dependence of 
the drift function and dispersion; hence, \cref{eq:SSDE} can be understood as a collection of 
individual \acp{sde} between which the systems switches via $Z(t)$.
For an accessible introduction to \acp{sde}, see \cite{sarkkaAppliedStochasticDifferential2019}.

\paragraph{The Hybrid Process.}
In the following, we refer to the continuous value $Y(t)$ as the \emph{state} and to
the discrete value $Z(t)$ as the \emph{mode} of a hybrid process.
The hybrid process $\{Z(t),Y(t)\}_{t\geq 0}$ is fully characterized by its time-point wise marginal density
$p(y,z,t):=\partial_{y_1}\cdots \partial_{y_n} \mathbb P(Y(t)\leq y,Z(t)=z)$, where ``$\leq$'' has to be interpreted element-wise. This density is given as the solution to the \ac{gfpe} \cite{pawula1967generalizations,altintan2019hybrid}
\begin{equation}
     \partial_t p(y, z, t )     = \mathcal{A} p(y, z, t ) ,
      \label{eq:generalized_fkp_prior}
\end{equation}
with initial condition $p(y, z, 0)=p_0(z) p_0(y,z)$, where $p_0(y,z):=\partial_{y_1}\cdots \partial_{y_n}\mathbb P(Y(0)\leq y\mid Z(0)=z)$
denotes the initial density of the $Y$-process and $p(z,0)=p_0(z)\,\, \forall z \in \mathcal{Z}$
the initial probability mass function of the $Z$-process.
The operator  $\mathcal{A}(\cdot)=\mathcal{F}(\cdot)+\mathcal{T}(\cdot)$ is given via
\begin{equation*}
    \begin{split}
             \mathcal{F} \phi(y,z,t) &:=- \sum_{i=1}^n \partial_{y_i} \left\lbrace f_i(y,z, t) \phi(y,z,t)\right\rbrace  +  \frac{1}{2}\sum_{i=1}^n \sum_{j=1}^n  \partial_{y_i}  \partial_{y_j} \lbrace 
      D_{ij}(y, z, t) \phi(y,z,t) \rbrace \\
           \mathcal{T}\phi(y,z,t) &:=\sum_{z^\prime \in \mathcal{Z}\setminus z} \Lambda(z',z,t) \phi(y,z',t)- \Lambda(z,t) \phi(y,z,t),
    \end{split}
\end{equation*}
for an arbitrary test function $\phi: \, \mathbb R^n \times \mathcal Z \times \mathbb R_{\geq 0}\rightarrow \mathbb R$.
We provide a detailed derivation in \cref{sec:app_derivation_hme}.
 %
%
%
%
%
As the discrete process $Z(t)$ is independent of $Y(t)$, we further obtain 
the dynamics of the marginal distribution $p(z,t):=\mathbb P(Z(t)=z)$ 
by integrating over the continuous variable $Y$ in the \ac{gfpe}. This recovers 
the traditional master equation~\cite{norrisMarkovChains1997}
\begin{align}
        \frac{\mathrm d}{\mathrm d t} p(z,t) 
        &=\sum_{z' \in \mathcal Z \setminus z} \Lambda(z', z, t)p(z',t)-\Lambda(z, t)p(z,t),
\label{eq:master_eq}
\end{align}
with $p(z,0)=p_0(z)\,\, \forall z \in \mathcal{Z}$; for details, see \cref{sec:app_exact_marg_z_process}.

Note that a general, analytical solution to the \ac{pde} \labelcref{eq:generalized_fkp_prior}
does not exist. Numerical solvers utilizing schemes such as the 
finite differences or finite element method suffer from the curse of dimensionality 
and can in principle only be applied to very low-dimensional state spaces \cite{grossmann2007numerical} .
Even in low dimensions however, solvers need to be adapted to the problem at 
hand and may struggle due to, e.g., slow step-size adaptation.
On the other hand, sampling trajectories from a hybrid process $\{Y(t), Z(t)\}$ is straightforward:
One can draw the process $Z(t)$ by utilizing the Doob-Gillespie algorithm \cite{doob1945markoff,gillespie1976general}. Given this trajectory, the 
\ac{ssde} $Y(t)\mid Z(t)$ can be simulated using, e.g., an Euler-Maruyama or stochastic Runge-Kutta method~\cite{kloeden1992stochastic}.

\paragraph{The Observation Process.}
Lastly, we denote with $\{X_i\}_{i \in \mathbb N}$ the countable set of observed data points at times $\{t_i\}_{i \in \mathbb N}$. The observations $X_i \in \mathcal X^l$ are generated as $X_i \sim p(x_i \mid y_i)$, where $p(x_i \mid y_i):=\partial_{x_{i1}}\cdots \partial_{x_{il}} \mathbb P(X_i \leq x_i \mid Y(t_i)=y_i)$, $i \in \mathbb N$, by conditioning on the diffusion process $Y(t)$. The observation space $\mathcal{X}$ can be either discrete, $\mathcal{X}\subseteq \mathbb N$, or continuous, $\mathcal{X}\subseteq \mathbb R$.
Note that in our model, a continuous-time description for the latent processes is assumed, while  the observations are recorded at discrete time points. It therefore belongs to the class of continuous-discrete models, which have a long history in the filtering community \cite{sarkkaAppliedStochasticDifferential2019,maybeck1979stochastic,daum1984}.
This type of description is of great practical relevance as data is often recorded at discrete
time points while the system of interest in fact evolves continuously in time, see, e.g., \cite{cassandras2009introduction}.

\subsection{Exact State Inference}
We now show how the exact posterior inference problem is solved in principle; the detailed derivations
can be found in \cref{sec:app_exact_inference}. 
The inference problem consists of finding the posterior hybrid process $\{Z(t), Y(t)\,\vert\, x_{[1,N]}$\},
where we condition on a finite set $x_{[1,N]} = \{x_1,\dots, x_N\}$ of $N$
observations obtained at time points $\{t_1,\dots, t_N\}$ in the
interval $[0,T]$.
The posterior process is fully specified by its marginal density
$
   p(y, z , t \mid x_{[1,N]}):=\partial_{y_1}\cdots \partial_{y_n} \mathbb P(Y(t)\leq y, Z(t)=z \mid X_1=x_1,\dots, X_N=x_N),
$
which is known as the \emph{smoothing distribution}.
The smoothing distribution is given by 
\begin{equation}
    p(y, z , t \mid x_{[1,N]})=C^{-1}(t) \alpha(y,z,t) \beta(y,z,t),
    \label{eq:exact_smoothing_def}
\end{equation} 
with the filtering density $\alpha(y,z,t):=\partial_{y_1}\cdots \partial_{y_n} \mathbb P(Y(t) \leq y ,Z(t)=z \mid X_1=x_1,\dots, X_k=x_k)$, 
the backward density $\beta(y, z, t):= \prod_{i=1}^l\prod_{j=k+1}^N\partial_{x_{{j}_{i}}} \mathbb P(X_{k+1}\leq x_{k+1},\dots,X_{N}\leq x_N\mid Y(t)=y , Z(t)=z)$
and a time-dependent normalizer $C(t)=\sum_z \int \alpha(y,z,t) \beta(y,z,t)\, \mathrm d y$, where $k=\max(k' \in \mathbb N \mid t_{k'} \leq t)$. The components $\alpha$, $\beta$ and $C$ are continuous-time analogs to the quantities of the forward-backward algorithm for discrete-time \acp{hmm} \cite{barber2012}.

It is a standard result for continuous-discrete filtering problems 
\cite{sarkkaAppliedStochasticDifferential2019} that the filtering distribution between 
observation time points follows the prior dynamics, 
$\partial_t \alpha(y, z, t )= \mathcal{A} \alpha(y, z, t )$,
with initial condition $\alpha(y, z, 0 )=p_0(z) p_0(y,z)$. At the observation times, 
it is reset as
$\alpha(y,z,t_i)= \tilde{C}_i^{-1} \alpha(y,z,t_i^-)p(x_i\mid y)$, with the normalizer
$\tilde{C}_i=\sum_z \int \alpha(y,z,t_i^-)p(x_i\mid y)\, \mathrm d y$ 
and $\alpha(y, z, t_i^-) := \lim_{h\searrow0}\alpha(y, z, t_i - h)$.
Similarly, the backward distribution between observations is given as the solution to another \ac{pde}~\cite{andersonSmoothingAlgorithmsNonlinear1983} 
\begin{equation}
\partial_t \beta(y, z, t ) =-{\mathcal A}^\dagger \beta(y, z, t),
    \label{eq:generalized_fkp_backward_distribution}
\end{equation} with end point condition $\beta(y,z,T)=1$ and adjoint operator ${\mathcal A}^\dagger$, see \cref{sec:app_derivation_hme}.
The reset conditions at observation times are given as $\beta(y,z,t_i^-)= \beta(y,z,t_i)p(x_i\mid y)$. 

By calculating the time derivative of \cref{eq:exact_smoothing_def}, it can be shown (see \cref{sec:app_derivation_smoothing}) that the smoothing distribution itself follows a \ac{gfpe}
\begin{equation}
    \partial_t p(y,z,t \mid x_{[1,N]}) = \tilde{\mathcal A} p(y,z,t \mid x_{[1,N]}), 
    \label{eq:generalized_fkp_posterior}
\end{equation}
with initial condition $p(y,z,0 \mid  x_{[1,N]}) \propto p_0(z) p_0(y, z) \beta(y,z,0)$.
The operator $\tilde{\mathcal A}$ contains the posterior drift function
$
    \tilde{f}_i(y,z,t)= f_i(y,z, t)+\sum_{j=1}^n D_{i j}(y,z,t) \partial_{y_j}\lbrace\log \beta(y,z,t)\rbrace,
$
the dispersion matrix 
$
    \tilde{D}(y,z,t)=D(y,z,t)
$
and the posterior rate function 
$
    \tilde{\Lambda}(z',z,t)=\Lambda(z',z,t)\frac{\beta(y,z,t)}{\beta(y,z',t)}.
$

%% file: sections/04_inference.tex
\section{Approximate Inference}
\label{sec:inference}
Since the smoothing distribution is governed by the \ac{gfpe} \labelcref{eq:generalized_fkp_posterior}, 
which depends on the solution of \cref{eq:generalized_fkp_backward_distribution}, 
the exact inference problem amounts to solving two \acp{pde}, which is computationally 
intractable already for toy systems. Similarly, a naïve posterior sampling scheme would still require solving the
backward \ac{pde} \cref{eq:generalized_fkp_backward_distribution} and hence suffers from
the same issue.
To address this challenge, we adopt a \ac{vi} approach: we aim to find
an approximate path measure $\mathbb Q_{Y,Z}$ that minimizes the path-wise \ac{kl} divergence
\begin{equation}\label{eq:KL}
    \KL\left(\mathbb Q_{Y,Z} \mid\mid \mathbb P_{Y,Z\mid X} \right)={\mathsf E}_{\mathbb Q_{Y,Z}}\left[\log \frac{\mathrm d \mathbb Q_{Y,Z}}{\mathrm d \mathbb P_{Y,Z\mid X} }\right],
\end{equation}
where $\frac{\mathrm d \mathbb Q_{Y,Z}}{\mathrm d \mathbb P_{Y,Z\mid X}}$ is the 
Radon-Nikodym derivative between $\mathbb Q_{Y,Z}$ and the 
exact posterior measure $\mathbb P_{Y,Z\mid X}$ over paths 
$Y_{[0,T]}:=\{Y(t)\}_{t \in [0,T]}$ and $Z_{[0,T]}:=\{Z(t)\}_{t \in [0,T]}$.
For details on the path-wise \ac{kl} divergence between stochastic processes, see, e.g., 
\cite{sutter2016,mitter2003variational,matthews2016sparse}.
It is a standard result for \ac{vi} methods that \cref{eq:KL} can be recast as~\cite{bleiVariationalInferenceReview2017,vanhandelFilteringStabilityRobustness2007}
\begin{equation}
     \KL(\mathbb Q_{Y, Z}\mid\mid \mathbb P_{Y, Z\mid X}) = \KL(\mathbb Q_{Y, Z}\mid\mid \mathbb P_{Y, Z})
     -\E_{\mathbb Q_{Y,Z}}[\ln p(x_{[1,N]}\mid y_{[0,T]})] + \log p(x_{[1,N]}),
\end{equation}
with the expected log-likelihood 
$\E_{\mathbb Q_{Y,Z}}[\ln p(x_{[1,N]}\mid y_{[0,T]})]=\sum_{i=1}^N \E_{\mathbb Q_{Y,Z}}[\ln p(x_{i}\mid y_{i})]$.
The minimization problem over \cref{eq:KL} can then be cast as a maximization 
problem over the \ac{elbo} \cite{bleiVariationalInferenceReview2017}
\begin{equation}
    \mathcal{L}[\mathbb Q_{Y, Z}]= \E_{\mathbb Q_{Y,Z}}[\ln p(x_{[1,N]}\mid y_{[0,T]})] -\KL(\mathbb Q_{Y, Z}\mid\mid \mathbb P_{Y, Z}),
    \label{eq:elbo}
\end{equation}
which does not include the computationally intractable marginal log-likelihood
$\log p(x_{[1,N]})$ and can hence be evaluated. 

Optimizing \cref{eq:elbo} requires explicitly computing the path-wise
\ac{kl} divergence $\KL(\mathbb Q_{Y, Z}\mid\mid \mathbb P_{Y, Z})$. 
For two hybrid processes of the same class obeying \cref{eq:generalized_fkp_prior}, 
this expression can formally be derived using Girsanov's theorem for
diffusion processes \cite{sottinen2008} 
and \acp{mjp} \cite{kipnis1998scaling}. A more intuitive derivation can however be 
carried out using a limiting argument, similar to \cite{opperVariationalInferenceMarkov2008}.
Assuming a constant, state- and mode-independent dispersion $D$ as done also in
\cite{archambeauVariationalInferenceDiffusion2008} and a drift $g(y, z, t)$ and rate function $\tilde{\Lambda}(z, z', t)$ pertaining
to the variational measure $\mathbb{Q}_{Y, Z}$,
the path-wise \ac{kl} divergence is obtained as
\begin{align}
   &\KL\left(\mathbb Q_{Y, Z} \mid\mid \mathbb P_{Y, Z}\right) = \KL(\mathbb Q^0_{Y, Z}\mid\mid \mathbb P^0_{Y, Z})+ \frac{1}{2}\int_0^T \E\left[\Vert(g(y, z, t) - f(y, z, t)\Vert_{D^{-1}}^2\right.\nonumber\\
   &{}\quad +\left.\sum_{z' \in \mathcal Z \setminus z} \left\lbrace \tilde{\Lambda}(z,z',t)\left(\ln \tilde{\Lambda}(z,z',t) -\ln\Lambda(z,z',t) \right)\right\rbrace-(\tilde{\Lambda}(z,t)-\Lambda(z,t))\right]\,\mathrm d t,
    \label{eq:kl_ssdes}
\end{align}
with the weighted norm $\Vert x \Vert^2_A:=x^\top A x$, the \ac{kl} of the initial 
distributions, $\KL(\mathbb Q^0_{Y, Z}\mid\mid \mathbb P^0_{Y, Z})$, and the expectation
is carried out with respect
to the variational time-point marginal $q(y,z,t):= \partial_{y_1}\cdots \partial_{y_n} \mathbb Q(Y(t)\leq y, Z(t)=z)$. We note that extensions to mode- and time-dependent $D = D(z, t)$
\cite{sottinen2008} or state-dependent $D=D(y)$ \cite{sutter2016} are also possible.
Additionally, in the absence of any coupling between $Z(t)$ and $Y(t)$, i.e., $f(y, z, t)=f(y, t)$,
\cref{eq:kl_ssdes} reduces to the sum of the known individual path-wise \ac{kl} divergences for
diffusion processes and \acp{mjp}.
The detailed derivation can be found in~\cref{sec:app_derivation_kl}.

\subsection{The Constrained Objective}
\label{sec:constrained_obj}
Since \cref{eq:elbo} is a mere reformulation, it is still optimized by the true,
intractable posterior distribution $\mathbb{P}_{Y, Z\mid X}$. 
To arrive at computationally tractable expressions, we restrict the class $\mathcal Q$ of 
admissible variational processes.
Making a structured mean-field ansatz, we approximate the exact joint posterior density
$p(y, z, t\mid x_{[1,N]})$ as
\begin{align}
    \begin{split}
    p(y, z, t\mid x_{[1,N]}) &=p(z,t \mid x_{[1,N]}) \cdot p(y, t \mid z, x_{[1,N]})\\
               &\approx q_Z(z, t)\cdot q_Y(y, t \mid z) = q(y, z, t),
    \end{split}
    \label{eq:approximation}
\end{align}
with $p(y, t \mid z, x_{[1,N]}):=\partial_{y_1}\cdots \partial_{y_n} \mathbb P(Y(t)\leq y \mid Z(t)=z, x_{[1,N]})$.
We approximate the exact conditional $p(y, t \mid z, x_{[1,N]})$, which in general does not
have a simple parametric form, by one fixed parametric expression per mode 
$q_Y(y, t\mid z) := \partial_{y_1}\cdots \partial_{y_n} \mathbb Q(Y(t) \leq y \mid \tilde{Z} = z)$ via
the introduction of the time-independent random variable $\tilde{Z}$. 
This results in a point-wise mixture distribution
$q(y, z, t)$ with weights $q_Z(z, t):=\mathbb Q(Z(t)=z)$ and mixture densities $q_Y(y, t \mid z)$. 
Note that this is similar in spirit to amortized inference techniques \cite{zhang2018}, because 
we utilize the same parametric form for all times $t$.

To ensure the mixture distribution structure \cref{eq:approximation} to hold at every time 
point $t$, we impose separate constraints on the dynamics of
$q_Z(z, t)$ and $q_Y(y, t\mid z)$. Firstly, we require the marginal $q_Z(z, t)$ to obey
a master equation
\begin{equation}\label{eq:constraint_meq}
     \frac{\mathrm d}{\mathrm d t} q_Z(z,t) = \sum_{z' \in \mathcal Z \setminus z} \tilde{\Lambda}(z',z, t)q_Z(z',t)-\tilde{\Lambda}(z,t)q_Z(z,t), \; \forall z \in \mathcal Z.
\end{equation}
This reproduces the structure of the exact posterior marginal $p(z, t \mid x_{[1,N]})$, which can
be seen by integrating out the continuous variable in the \ac{gfpe}
\labelcref{eq:generalized_fkp_posterior}, c.f. \cref{eq:master_eq}.
Secondly, we constrain the variational factor $q_{Y}(y, t\mid z)$ to follow a \ac{fpe} \cite{oksendal2003stochastic}
with linear variational drift $g(y, z, t) = A(z, t)y + b(z, t)$
for every mode $z$ individually
\begin{equation}\label{eq:mode_fpe}
    \partial_t q_{Y}(y, t\mid z) = - \sum_{i=1}^n \partial_{y_i} \left\lbrace g_i(y,z, t)
    q_{Y}(y, t\mid z)\right\rbrace +  \frac{1}{2}\sum_{i=1}^n \sum_{j=1}^n
    \partial_{y_i}  \partial_{y_j} \lbrace D_{ij} q_{Y}(y, t\mid z) \rbrace.
\end{equation}
\Cref{eq:mode_fpe} describes the marginal density of a classical \ac{sde}, which, under 
linear drift, is equivalent to a \ac{gp} \cite{rasmussenGPBook}.
This \ac{pde} is hence solved by a time-dependent Gaussian distribution $q_{Y}(y, t\mid z)=\mathcal N(y \mid \mu(z,t),\Sigma(z,t) )$ \cite{sarkkaAppliedStochasticDifferential2019},
where the dynamics of the parameters is described by two \acp{ode}
\begin{equation}\label{eq:moment_odes}
        \dot{\mu}(z, t) = A(z, t)\mu(z, t) + b(z, t) , \quad
        \dot{\Sigma}(z, t) = A(z, t)\Sigma(z, t) + \Sigma(z, t)A^\top(z,t) + D, \; \forall z \in \mathcal Z.
\end{equation}
Our approach hence amounts to a mixture of \acp{gp}: this approximation will be accurate
whenever the distribution $q_Z(z, t)$ is peaked at one $z \in \mathcal{Z}$. 
In this case, the \ac{gfpe} separates into a \ac{fpe} and a master 
equation for the processes $Y(t)\mid \tilde{Z}$ and $Z(t)$, respectively;
see \cref{sec:app_derivation_approximation} for details.
Accordingly, the approximation error over the whole interval $[0, T]$ will be small 
if the original system dynamics are linear in each mode and the modes are well
discernible. Since we are interested specifically in meta-stable systems
which, by definition, transition between qualitatively different regimes and
exhibit a separation of time scales between the intra-mode diffusive 
dynamics and the inter-mode transitions, we expect these criteria to be met reasonably well
for our systems of interest.

The constraints \labelcref{eq:constraint_meq,eq:moment_odes} can be included into the objective 
\cref{eq:elbo} via Lagrange multiplier functions, yielding an augmented objective, the \emph{Lagrangian}.
We define the multipliers $\lambda(z, t)$, $\Psi(z, t),\nu(z, t)$ for the variational mean $\mu(z, t)$ and covariance $\Sigma(z, t)$
and variational rates $\tilde{\Lambda}(z,z',t)$, respectively. 
Writing the \ac{elbo} as $\mathcal{L}[\mathbb Q_{Y, Z}] = \int_0^T\ell_{\mathbb Q}(t)\,\mathrm dt$, the full Lagrangian $L$ to be maximized reads
\begin{equation}
\begin{split}
        &L= \int_0^T \ell_{\mathbb Q}(t) + \sum_{z \in \mathcal{Z}}\left[\lambda^\top(z, t)\left( \dot{\mu}(z, t)  - \right. 
        \left(A(z, t)\mu(z, t) + b(z, t) \right)\right) +\tr\left\{\Psi^\top(z, t) \left(\dot{\Sigma}(z, t)\right. \right.\\
        &\left. \left. 
        -\left(A(z, t)\Sigma(z, t) + \Sigma(z, t)A^\top(z, t) + D\right)\right)\right\}+ \left.\nu(z,t) \left(\dot{q}_Z(z,t) - \sum_{z' \in \mathcal Z} \tilde{\Lambda}_{z'z}(t)q_Z(z',t) \right)\right] \,\mathrm d t,
\end{split}
    \label{eq:Lagrangian}
\end{equation}
where we used the shorthand $ \tilde{\Lambda}_{z'z}(t):= \tilde{\Lambda}(z,'z,t)$ for $z' \neq z$
and $\tilde{\Lambda}_{zz}(t):= -\tilde{\Lambda}(z,t)$ else. Note that the dependency of $L$ on
the variational measure $\mathbb Q_{Y, Z}$ is fully captured by the variational factors, 
$L = L[\mathbb Q_{Y, Z}] = L[q_Z, \mu, \Sigma]$.

\subsection{Optimizing the Variational Distributions}
The optimization problem consists in finding the optimal variational factors $q_Z^*, \mu^*, \Sigma^*$
and parameters $A^*, b^*, \tilde{\Lambda}^*, \phi^*$ maximizing \cref{eq:Lagrangian}, where 
$\phi$ summarizes the variational initial conditions. The optimal variational
factors have to fulfil the respective constraint equations \labelcref{eq:constraint_meq,eq:moment_odes} as well as
the \ac{el} equation $\frac{\mathrm d}{\mathrm dt}\partial_{\dot{q}} \ell = \partial_q \ell $,
where $L = \int_0^T \ell(t)\,\mathrm dt$ \cite{liberzonCalculusVariationsOptimal2012}. The latter is
giving rise to \acp{ode} for the Lagrange multiplier functions: Firstly, the \ac{el} equation with
respect to $q_Z(z, t)$ yields
\begin{equation}
        \dot{\nu}(z, t) =
    \partial_{q_Z(z, t)} \ell_{\mathbb Q} -\sum_{z' \in \mathcal Z \setminus z}\tilde{\Lambda}(z,z',t)\nu(z', t) + \tilde{\Lambda}(z,t)\nu(z, t) .
    \label{eq:MJP_lagrange_equation}
\end{equation}
Secondly, the \ac{el} equations hold separately for both Gaussian parameters $\mu(z, t), \Sigma(z, t)$.
We obtain
\begin{equation}
    \begin{split}
        \dot{\lambda}(z, t)&=\partial_{\mu(z,t)}  \ell_{\mathbb Q}- A^\top(z, t)\lambda(z, t),\\
        \dot{\Psi}(z, t)&= \partial_{\Sigma(z, t)}  \ell_{\mathbb Q} - A^\top(z, t)\Psi(z, t) 
        - \Psi(z, t)A(z, t).
    \end{split}
    \label{eq:L_multipliers_GPA}
\end{equation}
For a detailed derivation, including the explicit expressions for the gradients 
$\partial \ell_{\mathbb Q}$ for linear prior models $f(y, z, t) = A_p(z, t)y + b_p(z, t)$, see \cref{sec:app_derivation_optimzing_objective}.

The full optimization problem requires the Lagrange multiplier \acp{ode} 
\labelcref{eq:MJP_lagrange_equation,eq:L_multipliers_GPA} and the constraint
\cref{eq:constraint_meq,eq:moment_odes} to be solved jointly as a boundary-value problem 
with terminal conditions $\nu(\cdot, T), \lambda(\cdot, T), \Psi(\cdot, T)=0$ and
initial conditions on the distribution parameters \cite{liberzonCalculusVariationsOptimal2012}.
The variational parameters have to be optimized simultaneously.
A standard approach to this problem is an iterative forward-backward sweeping algorithm 
\cite{stengel1994,mcasey2012convergence}:
(i) solve the Lagrange multiplier \acp{ode} \cref{eq:MJP_lagrange_equation,eq:L_multipliers_GPA} backward in time, starting from the terminal conditions $\nu,\lambda,\Psi=0$. 
Next,
(ii) update the variational parameters acting on the constraints, in our case
$A, b, \tilde{\Lambda}, \phi$. Here, we employ a simple gradient
ascent scheme: for each $u(t)\in \{A(z, t), b(z, t), \tilde{\Lambda}(z,z',t), \phi\}$, we update
\begin{equation}
        u(t) \gets u(t) + \kappa(t) \cdot \partial_{u(t)}\ell,
    \label{eq:line_search}
\end{equation}
where we use a back-tracking line search \cite{bertsekas1997nonlinear} for the step size $\kappa(t)$, see \cref{sec:app_control_gradients}. Then, 
(iii) solve the constraint equations \labelcref{eq:constraint_meq,eq:moment_odes} forward in time, starting from initial conditions
$\phi = \{q_Z(z, 0) = q_Z^0(z), \mu(z, 0)=\mu^0(z), \Sigma(z, 0)= \Sigma^0(z)\}$. Finally,
(iv) repeat until convergence.

Note that our results generalize the findings of \cite{archambeauGaussianProcessApproximations2007,opperVariationalInferenceMarkov2008}: 
for $f(y, z, t) = f(y, t)$, $g(y, z, t) = g(y, t)$, i.e., in the absence of coupling 
between the $Y$- and $Z$-processes, the \ac{mjp} and diffusion contributions to the prior
\ac{kl} \cref{eq:kl_ssdes} separate and the derivative $\partial_{q_Z(z, t)} \ell_{\mathbb Q}$
reduces to the result of \cite{opperVariationalInferenceMarkov2008} between observations.
Furthermore, for the special case $\vert\mathcal{Z}\vert = 1$, our result recovers the conventional \ac{gp}
approximation \cite{archambeauGaussianProcessApproximations2007}.

\subsection{Parameter Learning}
To learn the model parameters, that is, the prior transition rate matrix $\Lambda$, 
the dispersion $D$, the prior initial conditions, the parameters of the drift function $f(y, z, t)$
and the parameters of the observation likelihood $p(x_i\mid y_i)$, we employ a \ac{vem} scheme
\cite{barber2012}. After converging onto variational distributions $q_Z(z, t)$ and $q_Y(y, t\mid z)$,
we perform a simple gradient ascent scheme on the Lagrangian $L$, \cref{eq:Lagrangian}, see \cref{sec:app_parameter_gradients}.
The complete optimization scheme is summarized in \cref{algorithm}, where we subsume all
model parameters under $\Theta$ for conciseness.

%% file: sections/05_experiments.tex
\section{Experiments}
\label{sec:experiments}
\subsection{Model Validation on Ground-Truth Data}

\begin{wrapfigure}{rb}{0.45\textwidth}
\centering
\begin{algorithm}[H]
\SetKwInOut{Input}{input}
\Input{observation data $\{t_i, x_i\}_{i=1, ..., N}$}
\BlankLine
Initialize $q_Z$, $\mu$, $\Sigma$, $A$, $b$, $\tilde{\Lambda}$, $\Theta$\\
\While{$\mathcal{L}$ not converged}{
    \While{$\mathcal{L}$ not converged}{
        Compute multiplier functions $\lambda$, $\Psi$, $\nu$ via \cref{eq:MJP_lagrange_equation,eq:L_multipliers_GPA}\\
        Update variational parameters $A$, $b$, $\tilde{\Lambda}$ via \cref{eq:line_search}\\
        Compute variational factors $\mu$, $\Sigma$, $q_Z$ via \cref{eq:moment_odes,eq:constraint_meq}\\
        Update lower bound $\mathcal{L}$
    }
    Update prior parameters $\Theta$ via gradient ascent\\
    Update lower bound $\mathcal{L}$
}
\caption{\acs{vi} for hybrid processes}
\label{algorithm}
\end{algorithm}
\end{wrapfigure}

We validate our method on synthetic data generated from a 1D, two-mode hybrid system
with observations corrupted by Gaussian noise, 
$p(x_i\mid y_i) = \mathcal{N}(x_i \mid y_i, \Sigma_{\mathrm{obs}})$, where the observation
times are drawn from a Poisson point process \cite{ethierMarkovProcessesCharacterization2005}.
Both mode dynamics are given by linear drift functions
\begin{equation}\label{eq:linear_drift}
    f(y, z, t) = \alpha_{z}(\beta_{z} - y),
\end{equation}
with set points $\beta_z$ and dynamics $\alpha_z > 0$, $z\in\mathcal{Z} = \{1, 2\}$. For $\vert \mathcal{Z}\vert  = 1$, this would recover the well-known 
Ornstein-Uhlenbeck process \cite{sarkkaAppliedStochasticDifferential2019}. 

\begin{figure}[b]

        \includegraphics[width=\columnwidth]{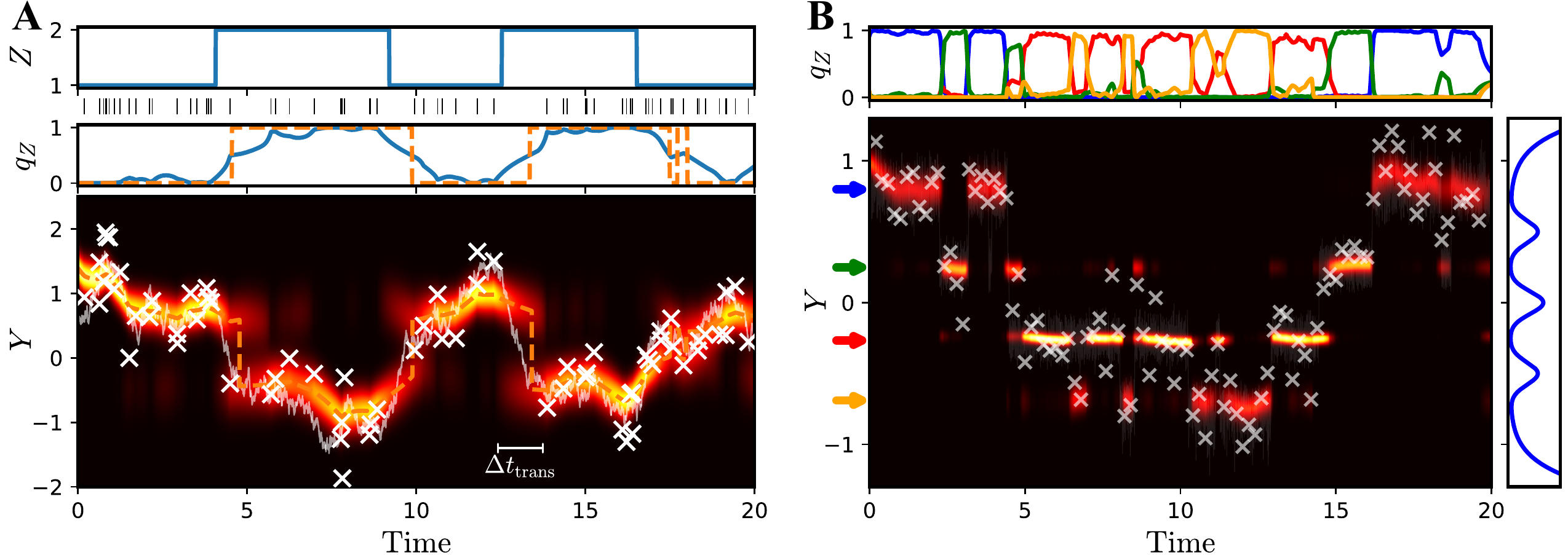}

    \caption{\textbf{A}: Model validation on ground-truth data from a 1D, two-mode hybrid process. Top: True discrete path $z(t)$ with observation times (vertical lines below). 
    Middle: inferred 
    marginal $q_Z(2, t)$ and $z^{\mathrm{MAP}}(t)$ (dashed). Bottom: true \ac{ssde} path (gray line), observations (crosses), the inferred marginal $q_Y(y, t)$ and $y^{\mathrm{MAP}}(t)$ (dashed).
    Brighter colors indicate higher probability density. \textbf{B}: Diffusion in a 1D four-well
    potential (shown right). Top: inferred marginals $q_Z(z, t)$. Bottom:
    true \ac{sde} and observations with the inferred $q_Y(y, t)$ and learned set points $\beta_z$ (arrows).}
    \label{fig:synSSDE}
\end{figure}

As shown in \cref{fig:synSSDE}~A, the inferred posterior distributions $q_Y(y, t)=\sum_{z\in \mathcal{Z}}q_Y(y, t\mid z)$ and 
$q_Z(z, t)$ both faithfully reconstruct the respective latent ground-truth trajectories.
This is also reflected by the \ac{map} paths $y^{\mathrm{MAP}}(t),z^{\mathrm{MAP}}(t)=\argmax_{y,z}q(y, z, t)$.
In regions around mode transitions, one can observe artifacts from the variational 
approximation as a mixture of \acp{gp}: in the time interval $\Delta t_{\mathrm{trans}}$
between the last observation before and the first observation after the ground-truth
mode transition at $t\approx 12.5$, the marginal $q_Y(y, t)$ does not exhibit a smooth 
transition across $Y=0$, but splits the probability density 
between the independently evolving Gaussian distributions $q_Y(y, t \mid z)$. This is not surprising,
as we have argued (c.f. \cref{sec:constrained_obj}) that our approximation will be accurate
in regions where $q_Z(z, t) \approx 1$, which is the case at the beginning and the end
of the interval $\Delta t_1$, but not in between. Since the relaxation onto the mode set points 
$\beta_z$ is fast compared to the mode remain times, these transition regions are short,
yielding a high approximation quality. 
As we pursue a generative modeling approach, we can verify this by sampling full trajectories
from the variational posterior. The empirical distribution over paths closely resembles the
latent continuous trajectory. We provide a plot of the sampling distribution in
\cref{sec:app_exp_synSSDE}. Furthermore, the model parameters are identified with high
accuracy: the learned set 
points, for instance, $\beta_1 = 0.72, \beta_2 = -0.41$, where the ground-truth 
values are $\pm 1$. The exhaustive 
list of both the learned and ground truth model parameters is also provided in 
\cref{sec:app_exp_synSSDE}. 

\subsection{Diffusions in Multi-Well Potentials}
In many real-world scenarios, discrete modes that drive continuous dynamics do not 
exist a priori, but continuous dynamics often exhibit regimes that are qualitatively 
different from one another. Transitions between different regimes typically occur on 
vastly longer time scales than the relaxation dynamics within each regime, as is 
observed, e.g., for the folding dynamics of complex biomolecules
\cite{sponerRNAStructuralDynamics2018}.
For such meta-stable systems in particular, probabilistic models explicitly assuming
the existence of discrete modes can aid understanding and enable targeted interventions
on the system. To demonstrate the capability of our model to yield sensible representations of 
distinct dynamic regimes, we apply it to latent It\^{o} diffusions
driven by 1D and 2D benchmark potentials widely used in computational biology
\cite{prinz2011,NIPS2018_7653,noeWu2013,wuNuske2017}
We model these system via hybrid processes with linear drift,
c.f. \cref{eq:linear_drift}, and Gaussian observation noise in both the 1D and 2D case, as before.
We assume a mode-dependent dispersion, $D=D(z)$.
The observation time points are regularly spaced and we fix the observation
covariance $\Sigma_{\mathrm{obs}}$.

\begin{figure}[t]

        \includegraphics[width=\columnwidth]{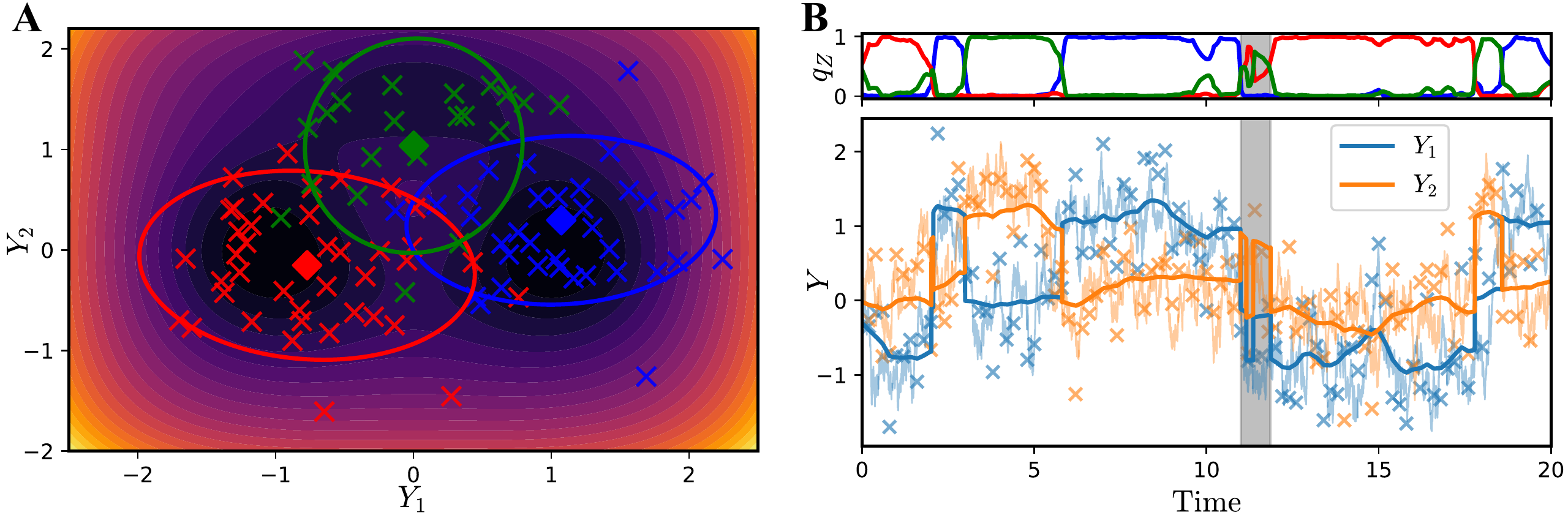}

    \caption{Diffusion in a three-well potential. \textbf{A}: Potential 
    landscape with the inferred set points $\beta_z$ (diamonds) and dispersions $D(z)$ (ellipses, $3\sigma$-region) with observations (crosses); colors according to $z^{\mathrm{MAP}}(t_i)$ for each observation $x_i$. \textbf{B}: Top: inferred marginals $q_Z(z, t)$. Bottom: components of $y^{\mathrm{MAP}}(t)$
    (thick lines), the ground-truth path (thin lines) and the observations (crosses). Shaded region: transition region with
    high ambiguity.}
    \label{fig:diffusions}
\end{figure}


In both the 1D and 2D case, the mode reconstructions accurately capture the global
transitions between distinct potential minima, see \cref{fig:synSSDE}~B and \cref{fig:diffusions}~B. 
We note that in the 1D example (\cref{fig:synSSDE}~B), the true latent continuous 
trajectory exhibits a particularly clear separation
of time scales between the inter- and intra-well dynamics, which is reflected in
sharp transitions in the mode reconstruction and accordingly particularly accurate results. 
On the other hand, as shown in \cref{fig:diffusions}, in the 2D case, more pronounced
transition regions exist, where it is not possible to unambiguously assign
the state at a given time $t$ to one of the three minima. The
posterior marginals $q_Z(z, t)$ sensibly capture this uncertainty, which
is also reflected in a high quality mode-assignment of the observed
data points $x_{[1, N]}$ as shown in \cref{fig:diffusions}~A.
Furthermore, the asymmetry of the learned mode-dependent dispersions accurately
reflects the topology of the underlying potential.
An overview over all parameters is given in \cref{sec:app_exp_diffusions}.

\subsection{Switching Ion Channel Data}
\begin{wrapfigure}[21]{rb}{0.45\textwidth}
\centering
    \includegraphics[width=0.45\columnwidth]{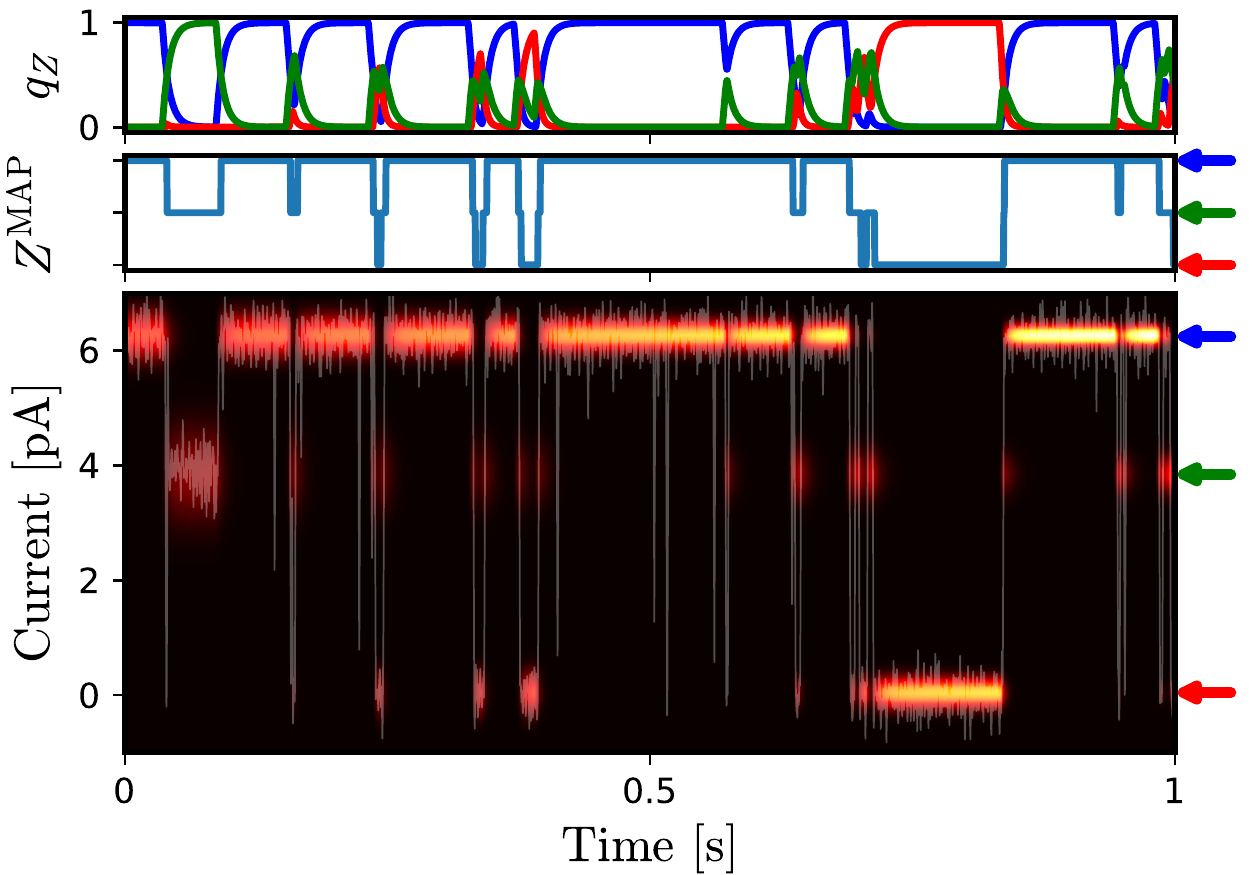}
    \caption{Switching behavior of viral ion channel $\text{Kcv}_{\text{MT325}}$. Top: marginals $q_Z(z, t)$ corresponding to three channel conformations.
    Middle: $z^{\mathrm{MAP}}(t)$.
    Bottom: data (white line), marginals $q_Y(y, t)$ and learned set points $\beta_z$ (arrows). Brighter colors indicate higher probability density.}
    \label{fig:ionchannel}
\end{wrapfigure}

We apply our method to a structural molecular biology problem: we aim to
identify the switching behavior of the viral ion channel $\text{Kcv}_{\text{MT325}}$ 
exhibiting three different channel conformations \cite{gazzarrini2006}. Different
conformations yield different ion permeabilities and hence different conductivities
which can be directly detected by applying a voltage across the cell membrane and
measuring the trans-membrane current. We model the conformation dynamics as discrete process $Z$
and the current, passing through an analog (continuous-time) filter and incurring amplifier noise,
as continuous process $Y$. The observations $X_i$ are sampled at a fixed rate and subject
to quantization errors from an analog-to-digital converter. The observation noise is a known
property of the used setup; we hence fix $\Sigma_{\mathrm{obs}}$.
We reconstruct a highly plausible switching behavior and filter out individual outliers, as depicted in \cref{fig:ionchannel}; see \cref{sec:app_exp_channels} for a list 
of all learned parameters and experimental details.
We note that very similar problem setups can be found, e.g., in nanopore sequencing
technologies \cite{jain2018nanopore}.\looseness-1

\subsection{Learning Complex Latent Continuous Dynamics}
\begin{figure}[b]
    \includegraphics[width=\columnwidth]{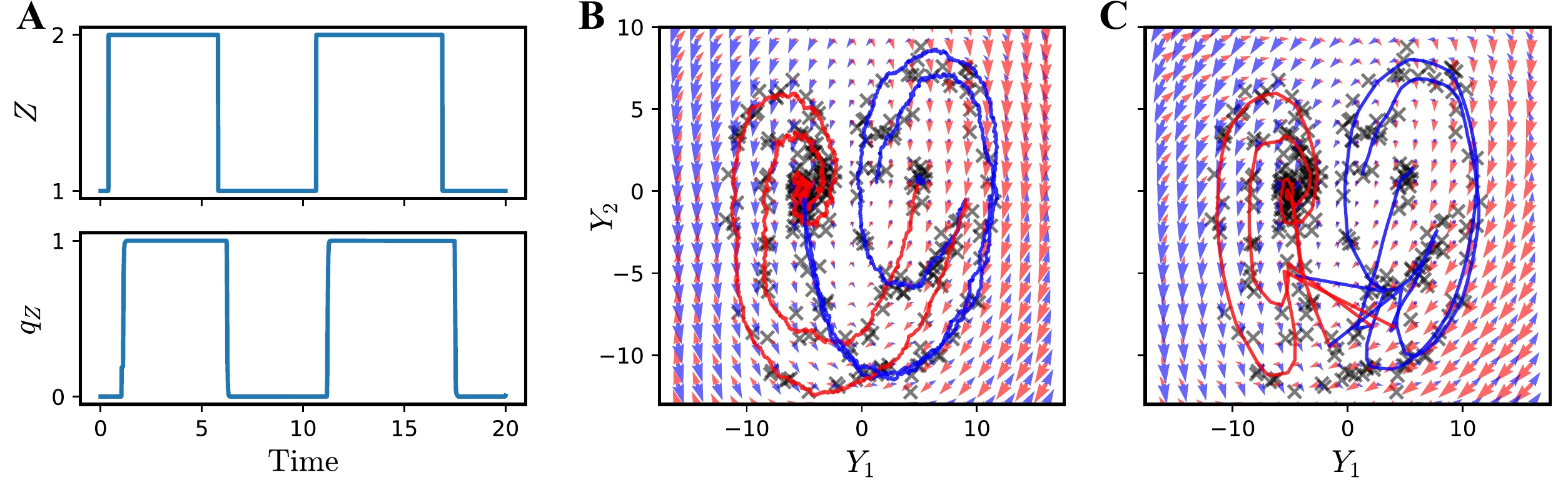}
    \caption{Inference of complex structured continuous dynamics.
    \textbf{A}: True mode path $z(t)$ (top) and inferred marginal $q_Z(2, t)$ (bottom). 
    \textbf{B}: True state path $y(t)$ (coloring according to associated mode; $Z=1$: blue, $Z=2$: red), observations (crosses) and true mode dynamics $f(y, z)$ (arrows) in the phase plane.
    \textbf{C}: Reconstructed \ac{map} path $y^{\mathrm{MAP}}(t)$ (coloring
    according to $z^{\mathrm{MAP}}(t)$) and reconstructed mode dynamics (arrows).
    }
    \label{fig:2D_swirls}
\end{figure}

After having demonstrated the applicability of our model to hybrid systems with 
different time scales for the discrete and continuous dynamics,
we lastly show that it also works well when this criterion is not met.
In areas such as automation and robotics, hybrid models are
ubiquitously used to, e.g., encode highly complex continuous movements
via a discrete set of movement primitives \cite{clever2017},
which are non-stationary processes.
To demonstrate that our method is able to reconstruct such complex
latent continuous dynamics, we employ a 2D version of \cref{eq:linear_drift} where the mode 
dynamics $f(y, z, t)=f(y, z)$ are given as two counter-rotating vector fields, see 
\cref{fig:2D_swirls} B. We fix the observation covariance, as we
can assume its value to be known and small compared to the system volatility
for applications such as robotics.

As shown in \cref{fig:2D_swirls} A and C, the mode and state reconstructions 
accurately recover the true paths. Accordingly, also the underlying mode dynamics
are correctly learned, exhibiting the counter-rotating behavior of the ground-truth model.
We provide a list of all parameters in \cref{sec:app_exp_swirls}.

%% file: sections/06_conclusion.tex
\section{Conclusion}

\label{sec:discussion}
We presented, to the best of our knowledge, the first variational inference framework for 
continuous time hybrid process models:
since the exact filtering and smoothing distributions are computationally intractable,
we proposed a variational approximation to the exact model.
The key assumption is that the true discrete posterior is peaked at any $z\in \mathcal{Z}$
for extended periods of time, allowing for a straightforward, easily interpretable
mixture of \acp{gp} to be used as approximation. We have evaluated our framework on
various benchmark tasks including real-world biological data and demonstrated its
ability to faithfully reconstruct complex latent dynamics and to learn the 
unknown system parameters, in particular in applications to meta-stable systems.
While we implemented parameter learning via point estimates, we aim
to extend this to a fully Bayesian framework in the future, enabling the integration
of prior domain knowledge and the associated uncertainty about the system at hand.
As many natural and engineered systems can be described as hybrid systems,
we think that extending the toolbox for inference will be 
of great utility for the analysis and control of such systems.


%% file: sections/07_acknowledgments.tex
\subsubsection*{Acknowledgments}
We thank Gerhard Thiel and Kerri Kukovetz for providing the ion channel voltage data and helpful discussions.
This work has been funded by the German Research Foundation~(DFG) as part of the project B4 within the Collaborative Research Center~(CRC) 1053 -- MAKI, and by the European Research Council (ERC) within the CONSYN project, grant agreement number 773196. 

%% file: appendix/A0_header.tex
\title{Variational Inference for Continuous-Time\\ Switching Dynamical Systems\\\normalfont{--- Supplementary Material ---}}
\maketitlenew

%% file: appendix/A1_appendix.tex
\section{Derivations}
\label{sec:app_derivations}
For the derivations we will use the notation 
\begin{equation*}
    p(x):=\partial_x \mathbb P(X\leq x)
\end{equation*}
for the density function of a random variable $X$. For the conditional density of a random variable $X$ and a realization $y$ of a random variable $Y$, we write 
\begin{equation*}
    p(x\mid y):=\partial_x \mathbb P(X\leq x \mid Y=y).
\end{equation*}
 For time-dependent densities with continuous random variables $Y(t)$ and discrete random variables $Z(t)$, we use the time-point marginal density 
 \begin{equation*}
     p(y,z,t):= \partial_y \mathbb P(Y(t)\leq y,Z(t)=z),
 \end{equation*}
time-point joint density 
\begin{equation*}
    p(y,z,t,y',z',t')=  \partial_y \partial_{y'} \mathbb P(Y(t)\leq y,Z(t)=z,Y(t')\leq y',Z(t')=z')
\end{equation*} and time-point conditional density 
\begin{equation*}
    p(y,z,t \mid y',z',t') =  \partial_y \mathbb P(Y(t)\leq y,Z(t)=z \mid Y(t')=y',Z(t')=z').
\end{equation*}
The latter also applies to conditional densities with multiple time points in 
the conditioning set,
\begin{equation*}
\begin{split}
        &p(y,z,t\mid y',z',t', y'', z'', t'')\\
        &{}:=\partial_y \mathbb P(Y(t)\leq y ,Z(t)=z\mid Y(t')=y',Z(t')=z',Y(t'')=y'',Z(t'')=z'').
\end{split}
\end{equation*}
If it is clear from the context, we will mostly use the favorable uncluttered notation.

\subsection{Derivation of the Hybrid Master Equation}
\label{sec:app_derivation_hme}
To derive the \ac{gfpe} conditioned on an arbitrary set $\mathcal{X}$, e.g., the set of
initial conditions $\mathcal{X} = \{Z(0)=z_0, Y(0)=y_0\}$, we assume for 
simplicity that $Z(t) \in \mathcal{Z} \subseteq \mathbb{N}$ and 
$Y(t) \in \mathcal{Y} \subseteq \mathbb{R}$. The multivariate case 
$\mathcal{Y} \subseteq \mathbb{R}^n$ is be derived analogously.

Following \cite{pawula1967generalizations}, we use the
rule of total probability on the density $p(y,z,t+h \mid \mathcal{X})$ for some $h > 0$:
\begin{align*}
    p(y, z, t + h \mid \mathcal{X})&= \sum_{z'}\int_\mathcal{Y} p(y, z, t+h \mid y', z', t, \mathcal{X})p(y', z', t\mid \mathcal{X})\,\mathrm dy'\\
    &= \sum_{z'}\int_\mathcal{Y} p(y, t+h \mid z, t+h, y', z', t, \mathcal{X}) p(z, t + h\mid y', z', t, \mathcal{X}) \\
    &\qquad\qquad\qquad \cdot p(y', z',t\mid \mathcal{X})\,\mathrm dy'.
\end{align*}

We expand, with $\lim_{h \rightarrow 0} \frac{1}{h} p(z, t+h \mid y', z', t, \mathcal{X}):= \Lambda_{z'z}^{y'}(t)$ and $\lim_{h \rightarrow 0}\frac{o(h)}{h}=0$,
\begin{align}
    p(z, t+h\mid y', z', t, \mathcal{X}) = \delta_{z'z} + \Lambda_{z'z}^{y'}(t)h + o(h)\nonumber
\end{align}
with the Kronecker delta $\delta_{z'z} = 1$ if $z'=z$ and 0 otherwise. Since we aim to 
take the limit $h \rightarrow 0$ at the end, we omit terms of 
$o\left(h\right)$. Inserting the expansion into the above expression, we obtain 
\begin{align}
    &p(y, z, t+h \mid \mathcal{X}) \nonumber\\
        &= \sum_{z'}\int_\mathcal{Y} p(y, t+h \mid z, t+h, y', z', t, \mathcal{X}) \left(\delta_{z'z} + \Lambda_{z'z}^{y'}(t)h\right) p(y', z', t\mid \mathcal{X})\,\mathrm dy'\nonumber\\
        \begin{split}
        &= \int_\mathcal{Y} p(y, t+h \mid z, t+h,y', z, t, \mathcal{X})p(y', z, t \mid \mathcal{X})\,\mathrm dy'\\
        &\qquad + \sum_{z'}\int_\mathcal{Y}p(y, t+h \mid z, t+h, y', z', t, \mathcal{X})\Lambda_{z'z}^{y'}(t)h p(y', z', t \mid \mathcal{X})\, \mathrm dy'\label{eq:joint_2}
        \end{split}
\end{align}
The density function $p(y, t+h \mid z, t+h, y', z', t, \mathcal{X})$
can be written in terms of its characteristic function $\psi(\nu, t +h\mid y', t, \mathcal{X}) = 
\E[e^{i\nu(Y(t + h) - Y(t))}\mid Z(t+h)=z, Y(t)=y', Z(t)=z', \mathcal{X}]$, 
\begin{align}\label{eq:characteristic_function}
     &p(y, t + h \mid z, t+h, y', z', t, \mathcal{X})\nonumber\\
     & = \frac{1}{2\pi}\int_\mathbb{R} e^{-i\nu (y-y')} \psi(\nu, t + h\mid y', t, \mathcal{X})\, \mathrm d\nu\nonumber\\
         \begin{split}
              & = \frac{1}{2\pi}\int_\mathbb{R} e^{-i\nu (y-y')} \\
     &\qquad\qquad \cdot \sum_{n=0}^\infty\frac{(i\nu)^n}{n!}\E[(Y(t + h) - Y(t))^n \mid Z(t+h)=z, Y(t)=y', Z(t)=z', \mathcal{X}]\,\mathrm d\nu,
         \end{split}
\end{align}
where we expressed the characteristic function via its Taylor series around $\nu=0$.
We insert this representation into \cref{eq:joint_2} and make use of the identity 
(which only holds under the integral)
\begin{equation*}
    \partial^{(n)}_{y} \delta(y - y') = \frac{1}{2\pi}(-i\nu)^n\int_\mathbb{R}e^{-i\nu(y-y')}\,\mathrm d\nu,
\end{equation*}
with $\partial^{(0)}_{y} \delta(y - y'):=  \delta(y - y')$, yielding
\begin{align*}
    &p(y, z, t+h \mid \mathcal{X}) \\
        &= \int_\mathcal{Y} p(y, t+h \mid z, t+h, y', z, t, \mathcal{X})p(y', z, t\mid X)\,\mathrm dy'\\
        &\qquad + h\cdot \sum_{z'}\int_\mathcal{Y}p(y, t+h \mid z, t+h, y', z', t, \mathcal{X})\Lambda_{z'z}^{y'}(t) p(y', z', t \mid \mathcal{X}) \,\mathrm dy'\\
        &= \int_\mathcal{Y} \sum_{n=0}^\infty \frac{(-1)^n}{n!} \partial^{(n)}_y\delta(y-y')\E[(Y(t + h) - Y(t))^n \mid Z(t+h)=z,\\
        &\qquad\qquad\qquad\qquad\qquad\qquad\qquad Y(t)=y', Z(t)=z, \mathcal{X}] p(y', z, t\mid \mathcal{X})\,\mathrm dy'\\
        &\qquad + h\cdot\sum_{z'}\int_\mathcal{Y}\sum_{n=0}^\infty \frac{(-1)^n}{n!} \partial^{(n)}_y\delta(y-y')\E[(Y(t + h) - Y(t))^n \mid Z(t+h)=z,\\
        &\qquad\qquad\qquad\qquad\qquad\qquad\qquad\qquad Y(t)=y', Z(t)=z', \mathcal{X}] \Lambda_{z'z}^{y'}(t)p(y', z', t \mid \mathcal{X})\, \mathrm dy'\\
        &= \sum_{n=0}^\infty \frac{(-1)^n}{n!} \partial^{(n)}_y\E[(Y(t + h) - Y(t))^n \mid Z(t+h)=z, Y(t)=y, Z(t)=z, \mathcal{X}]p(y, z, t\mid \mathcal{X})\\
        &\qquad + h\cdot\sum_{z'}\sum_{n=0}^\infty \frac{(-1)^n}{n!} \partial^{(n)}_y\E[(Y(t + h) - Y(t))^n \mid Z(t+h)=z, Y(t)=y, Z(t)=z', \mathcal{X}] \\
        &\qquad\qquad\qquad\qquad\qquad\qquad\qquad\qquad  \cdot\Lambda_{z'z}^{y}(t) p(y, z' , t\mid \mathcal{X}).
\end{align*}
We again apply a Taylor expansion and omit part of the conditioning set for brevity,
\begin{align}
    \E[(Y(t + h) - Y(t))^n\vert Y(t)=y] &= \sum_{m=0}^\infty \frac{h^m}{m!}\partial_\tau^{(m)}\E[(Y(t + \tau) - Y(t))^n\vert Y(t)=y]\vert_{\tau=0}\nonumber\\
    & = 1 + o\left(1\right)\label{eq:expectation_expansion}.
\end{align}
Inserting this and omitting terms $o\left(h\right)$, we obtain
\begin{align*}
    &p(y, z, t+h \mid \mathcal{X}) \\
        &= \sum_{n=0}^\infty \frac{(-1)^n}{n!} \partial^{(n)}_y\E[(Y(t + h) - Y(t))^n\vert Z(t+h)=z, Y(t)=y, Z(t)=z, \mathcal{X}]p(y, z, t\mid \mathcal{X})\\
        &\qquad + h\cdot\sum_{z'} \Lambda_{z'z}^{y}(t) p(y, z', t \mid \mathcal{X})\\
        &= p(y, z, t \mid \mathcal{X}) + \\
        &\qquad\sum_{n=1}^\infty \frac{(-1)^n}{n!} \partial^{(n)}_y\E[(Y(t + h) - Y(t))^n\vert Z(t+h)=z, Y(t)=y, Z(t)=z, \mathcal{X}]p(y, z, t\mid \mathcal{X})\\
        &\qquad + h\cdot\sum_{z'} \Lambda_{z'z}^{y}(t) p(y, z', t \mid \mathcal{X})
\end{align*}
Substracting $p(y, z, t \mid \mathcal{X})$ from both sides, dividing by $h$ and 
taking the limit $h \rightarrow 0$ yields
\begin{align}
    \partial_t p(y, z, t\mid \mathcal{X}) &= \lim_{h \rightarrow 0}\frac{p(y, z, t+h \mid \mathcal{X}) - p(y, z, t\mid \mathcal{X})}{h}\nonumber\\
    \begin{split}
    &= \sum_{n=1}^\infty \frac{(-1)^n}{n!} \partial^{(n)}_y \lbrace \Gamma_{nyz}p(y, z, t\mid \mathcal{X})\rbrace + \sum_{z'}\Lambda_{z'z}^y(t) p(y, z', t \mid \mathcal{X})
    \end{split}
\end{align}
with 
\begin{align}
\begin{split}
    \Gamma_{nyz} &= \lim_{h \rightarrow 0} \frac{1}{h}\E[(Y(t + h) - Y(t))^n\vert Z(t+h)=z, Y(t)=y, Z(t)=z, \mathcal{X}]\\
    \Lambda_{z'z}^y &= \lim_{h \rightarrow 0} \frac{1}{h} p(z, t+h \mid z', y', t, \mathcal{X}) - \delta_{z'z}.
\end{split}
\end{align}
As $Y(t)$ follows the \ac{ssde} 
\begin{equation*}
    \mathrm d Y(t)=f(Y(t),Z(t),t)\, \mathrm d t+Q(Y(t),Z(t),t)\, \mathrm d W(t),
\end{equation*}
we can compute the conditional moments $\Gamma_{nyz}$ in closed form. Conditioned on the discrete process remaining constant in a small time interval, $Z_{[t, t+h]}=z$,
the above \ac{ssde} can be treated as a conventional, $Z$-independent It\^o \ac{sde}.
For small $h$, we can hence utilize the usual Euler-Maruyama approximation~\cite{sarkkaAppliedStochasticDifferential2019},
$$
Y(t+h) \vert Z(t+h)=z, Z(t) = z, Y(t)=y \sim \NDis(y + f(y, z, t)h, D(y, z, t)h).
$$
Consequently,
\begin{align*}
    &\E[(Y(t + h) - Y(t))^n\vert Z(t+h)=z, Y(t)=y, Z(t)=z, \mathcal{X}]\nonumber \\
    &\qquad\qquad= \int (y' - y)^n \NDis(y + f(y, z, t)h, D(y, z, t)h)\,\mathrm dy'
\end{align*}
and the first two conditional moments are the usual Gaussian moments
\begin{equation*}
    \Gamma_{nyz}=\begin{cases}
        f(y,z,t) &\text{if } n=1\\
        \frac{1}{2}Q(y,z,t)Q^\top(y,z,t)=\frac{1}{2}D(y,z,t) &\text{if } n=2.\\
    \end{cases}
\end{equation*} 
As shown in \cite{pawula1967generalizations}, if $\Gamma_{nyz} = 0 $ for some even $n$,
$\Gamma_{nyz} = 0\,\forall n\geq 0$. It is straightforward to show, e.g., that 
$\Gamma_{nyz} = 0 $ for $n=4$, so all other conditional moments vanish.
Hence, we can (for arbitrary $\mathcal{Y} \subseteq \mathbb R ^n$) define the \ac{pde}
\begin{equation*}
    \partial_t p(y, z, t\mid \mathcal{X}) =\mathcal A p(y, z, t\mid \mathcal{X})
\end{equation*}
using the operator $\mathcal{A}(\cdot)=\mathcal{F}(\cdot)+\mathcal{T}(\cdot)$ as
\begin{equation*}
    \begin{split}
             \mathcal{F} p(y, z, t\mid \mathcal{X}) &=- \sum_{i=1}^n \partial_{y_i} \left\lbrace f_i(y,z, t) p(y, z, t\mid \mathcal{X})\right\rbrace\\
             &\qquad+  \frac{1}{2}\sum_{i=1}^n \sum_{j=1}^n  \partial_{y_i}  \partial_{y_j} \lbrace 
      D_{ij}(y, z, t) p(y, z, t\mid \mathcal{X}) \rbrace, \\
           \mathcal{T} p(y, z, t\mid \mathcal{X}) &=\sum_{z^\prime \in \mathcal{Z}\setminus z} \Lambda(z',z,t) p(y, z', t\mid \mathcal{X})- \Lambda(z,t) p(y, z, t\mid \mathcal{X}).
    \end{split}
\end{equation*}

In the same vein as the above derivation, using the Kolmogorov backward equation 
\begin{equation*}
    p(\mathcal X\mid y,z,t-h)=\sum_{z' \in \mathcal{Z}} \int p(y', z', t \mid y, z, t-h) p(\mathcal X\mid y',z',t) \,\mathrm d y',
\end{equation*}
we can find another \ac{pde} for the density $p(\mathcal X\mid y,z,t)$.
This yields the backward equation $\partial_t p(\mathcal X\mid y,z,t)= -\mathcal A^\dagger p(\mathcal X\mid y,z,t)$, with the adjoint operator ${\mathcal A}^\dagger (\cdot)=\mathcal{F}^\dagger(\cdot)+\mathcal{T}^\dagger(\cdot)$:
\begin{equation*}
    \begin{split}
             \mathcal{F}^\dagger p(\mathcal X\mid y,z,t) &= \sum_{i=1}^n f_i(y,z, t) \partial_{y_i} p(\mathcal X\mid y,z,t)+  \frac{1}{2}\sum_{i=1}^n \sum_{j=1}^n  D_{ij}(y, z, t) \partial_{y_i}  \partial_{y_j} 
     p(\mathcal X\mid y,z,t), \\
            \mathcal{T}^\dagger p(\mathcal X\mid y,z,t) &= \sum_{z^\prime \in \mathcal{Z}\setminus z} \Lambda(z,z',t)p(\mathcal X\mid y,z,t)- \Lambda(z,t)p(\mathcal X\mid y,z,t).
    \end{split}
\end{equation*}
The operator $\mathcal A^\dagger$ is adjoint to the operator $\mathcal A$, with respect to the inner product $\langle p,\phi \rangle:= \sum_z \int p(y,z,t) \phi(y,z,t) \, \mathrm d y$, i.e. 
\begin{equation*}
    \langle \mathcal A p,\phi \rangle=\langle p, \mathcal A^\dagger \phi \rangle
\end{equation*}
for an arbitrary test function $\phi$.

\subsubsection{Exact Marginal \texorpdfstring{$Z$}{Z}-Process} 
\label{sec:app_exact_marg_z_process}
We here show that integrating out the continuous variable $y$ from the \ac{gfpe} yields
the traditional master equation. The full \ac{gfpe} reads
\begin{align*}
     \partial_t p(y, z, t ) &= \mathcal{A} p(y, z, t )\nonumber\\
     \begin{split}
             &=- \sum_{i=1}^n \partial_{y_i} \left\lbrace f_i(y,z, t) p(y, z, t )\right\rbrace  +  \frac{1}{2}\sum_{i=1}^n \sum_{j=1}^n  \partial_{y_i}  \partial_{y_j} \lbrace 
      D_{ij}(y, z, t) p(y, z, t ) \rbrace \\
            &\qquad\qquad +\sum_{z^\prime \in \mathcal{Z}\setminus z} \Lambda(z',z,t) p(y, z', t )- \Lambda(z,t) p(y, z, t ).\label{eq:app_gfpe_zmarg}
    \end{split}
\end{align*}
Using Leibniz' theorem, we have 
\begin{align*}
    \int_\mathcal{Y}\partial_t p(y, z, t )\mathrm dy &= \partial_t \int_{\mathcal{Y}}p(y, z, t)\mathrm dy \nonumber \\
    &= \partial_t p(z, t).
\end{align*}
Accordingly, we have
\begin{equation}
    \begin{split}
          \partial_t p(z, t) =& 
           \underbrace{\int_\mathcal{Y} \left( - \sum_{i=1}^n \partial_{y_i} \left\lbrace f_i(y,z, t) p(y, z, t )\right\rbrace  +  \frac{1}{2}\sum_{i=1}^n \sum_{j=1}^n  \partial_{y_i}  \partial_{y_j} \lbrace 
      D_{ij}(y, z, t) p(y, z, t ) \rbrace\right) \, \mathrm d y}_{=0}
    \\
    &{}+ \underbrace{\int_{\mathcal{Y}}\sum_{z^\prime \in \mathcal{Z}\setminus z} \Lambda(z',z,t) p(y, z', t )- \Lambda(z,t) p(y, z, t )\, \mathrm d y}_{=\sum_{z^\prime \in \mathcal{Z}\setminus z} \Lambda(z',z,t) p(z', t )- \Lambda(z,t) p(z, t )}
    \end{split}
\end{equation}
and the first integral has to vanish because of the Gauss divergence theorem and $p(y, z, t) \overset{y\rightarrow 0}{\longrightarrow} 0$.
Hence, 
\begin{equation*}
    \partial_t p(z, t)=\sum_{z'\in \mathcal{Z}\setminus z}\Lambda(z', z, t)p(z', t) -\Lambda(z, t)p(z, t).
\end{equation*}
\subsection{Exact Posterior Inference}
\label{sec:app_exact_inference}

Here, we show how to calculate the quantities related to the smoothing density $p(y, z , t \mid x_{[1,N]}):=\partial_{y_1}\cdots \partial_{y_n} \mathbb P(Y(t)\leq y, Z(t)=z \mid X_1=x_1,\dots, X_N=x_N)$.
Using $k=\max(k' \in \mathbb N \mid t_{k'} \leq t)$ we can be express the smoothing density as 
\begin{align*}
    p(y, z , t \mid x_{[1,N]})&=\frac{p(y,z,t,x_1,\dots,x_k,x_{k+1},\dots,x_N)}{p(x_1,\dots,x_k,x_{k+1},\dots,x_N)}\\
    &=\frac{p(x_{k+1},\dots,x_N \mid x_1,\dots, x_k,y,z,t)}{p(x_{k+1},\dots,x_N\mid x_1,\dots,x_k)} \frac{p(y,z,t,x_1,\dots,x_k)}{p(x_1,\dots,x_k)}\\
    &=\frac{p(x_{k+1},\dots,x_N\mid y,z,t)}{p(x_{k+1},\dots,x_N\mid x_1,\dots,x_k)}p(y,z,t\mid x_1,\dots,x_k)\\
    &=C^{-1}(t) \alpha(y,z,t) \beta(y,z,t),
\end{align*}
with the filtering density $\alpha(y,z,t)=p(y,z,t\mid x_1,\dots,x_k)$, the backward density
$\beta(y,z,t)=p(x_{k+1},\dots,x_N\mid y,z,t)$  and a time-dependent normalizer
$C(t)=\sum_z \int \alpha(y,z,t) \beta(y,z,t)\, \mathrm d y$.

\subsubsection{Calculation of the Filtering Distribution}
\label{sec:app_filtering}
The filtering distribution is defined as
\begin{equation*}
     \alpha(y,z,t):= p(y,z,t \mid x_1,\dots,x_k ),
\end{equation*}
with density $p(y,z,t \mid x_1,\dots,x_k ):=\partial_{y_1} \cdots \partial_{y_n} \mathbb P(Y(t) \leq y,Z(t)=z \mid  X_1=x_1,\dots,X_k=x_k)$ and $k=\max(k' \in \mathbb N \mid t_{k'} \leq t)$.

\paragraph{The Filtering Distribution Between Observations.}
Consider the case where there is no observation in the interval $[t,t+h]$, $h>0$.

We compute
\begin{align*}
    \alpha(y,z,t+h)&=p(y,z,t+h \mid x_1,\dots,x_k)\\
    &=\sum_{z' \in \mathcal{Z}} \int  p(y, z, t+h, y',z',t \mid x_1,\dots,x_k) \,\mathrm d y'\\
    &=\sum_{z' \in \mathcal{Z}} \int p(y, z, t+h \mid y',z',t, x_1,\dots,x_k) p(y', z', t\mid x_1,\dots,x_k) \,\mathrm d y'. 
\end{align*}

As there are no observations in the interval $[t,t+h]$, we have
\begin{equation*}
    p(y, z, t+h \mid y',z', t, x_1, \dots,x_k)= p(y, z ,t+h \mid y',z', t).
\end{equation*}
This is true since the conditional process $\{Y(t+h),Z(t+h)\}$ given $\{Y(t), Z(t)\}$ is independent of $\{X_1,\dots,X_k\}$.
Hence, we have
\begin{equation*}
    \alpha(y,z,t+h)=\sum_{z' \in \mathcal{Z}} \int p(y, z ,t+h \mid y',z', t)\alpha(y',z',t))\,\mathrm{d} y'. 
\end{equation*}
This is the (forward) Chapman-Kolmogorov equation \cite{pawula1967generalizations} for the filtering process
$\{Y(t),Z(t) \mid x_1,\dots, x_k\}$ , with transition distribution $p(y, z ,t+h \mid y',z', t)$,
which is the transition distribution of the prior dynamics. Hence, between observations $\alpha(y,z,t)$ follows the \ac{gfpe}
\begin{equation*}
    \partial_t \alpha(y,z,t)=\mathcal A \alpha(y,z,t),
\end{equation*}
as derived in \cref{sec:app_derivation_hme}.

\paragraph{The Filtering Distribution at Observation Time Points.}
Here, we calculate the filtering distribution at the observation time points $\{t_i\}_{i \in 1,\dots, N}$.
\begin{align*}
    \alpha(y,z,t_i)&=p(y,z,t_i \mid x_1,\dots,x_i)\\
    &=\frac{p(y, z ,t_i, x_1, \dots, x_i)}{p(x_1, \dots, x_i)}\\
     &{}=\frac{p(x_i\mid y, z, t_i , x_1, \dots, x_{i-1}) p(y, z, t_i , x_1, \dots, x_{i-1})}{p(x_1, \dots, x_i)}\\
     &{}=\frac{p(x_i\mid y, z, t_i , x_1, \dots, x_{i-1})p(y, z,t_i \mid x_1, \dots, x_{i-1})p(x_1, \dots, x_{i-1})}{p(x_1, \dots, x_i) } \\
      &{}= \frac{p(x_i \mid y) \alpha(y,z,t_i^-)}{\tilde{C}_i}
\end{align*}
and $\tilde{C}_i=\frac{p(x_1, \dots, x_i)}{p(x_1, \dots, x_{i-1})}=\sum_{z \in \mathcal Z} \int p(x_i \mid y) \alpha(y,z,t_i^-) \,\mathrm d y$.

\subsubsection{Calculation of the Backward Distribution}
The backward distribution is defined as
\begin{equation*}
     \beta(y,z,t):= p(x_{k+1},\dots,x_N \mid y,z,t ),
\end{equation*}
with density $p(x_{k+1},\dots,x_N \mid y,z,t ):=\partial_{x_{k+1}}\cdots \partial_{x_{N}} \mathbb P(X_{n+1} \leq x_{n+1},\dots,X_{N}\leq x_N \mid Z(t)=y,Z(t)=z)$ and $k=\max(k' \in \mathbb N \mid t_{k'} \leq t)$

\paragraph{The Backward Distribution Between Observations.}
Consider again first the case where there is no observation in the interval $[t-h,t]$, $h>0$.
\begin{align*}
    \beta(y,z,t-h)&=p(x_{k+1},\dots, x_N \mid y,z,t-h)\\
    &=\sum_{z' \in \mathcal{Z}} \int p(x_{k+1},\dots, x_N, y', z',t \mid  y,z,t-h)\, \mathrm d y'\\
    &=\sum_{z' \in \mathcal{Z}} \int p(y', z',t \mid y, z,t-h) p(x_{k+1},\dots, x_N \mid  y', z',t, y, z,t-h)\, \mathrm d y'.
\end{align*}
As there are no observations in the interval $[t-h,t]$,
\begin{align*}
    &p(x_{k+1},\dots, x_N \mid  y', z', t, y, z,t-h)=p(x_{k+1},\dots, x_N \mid y',z',t)=\beta(y',z',t)
\end{align*}
as the process $\{x_{k+1},\dots, x_N \mid Y(t), Z(t) \}$ is independent of $\{Y(t-h),Z(t-h)\}$. Hence,

\begin{equation*}
      \beta(y,z,t-h)=\sum_{z' \in \mathcal{Z}} \int p(y', z', t \mid y, z, t-h) \beta(y',z',t) \,\mathrm d y'.
\end{equation*}

This is the (backward) Chapman-Kolmogorov equation \cite{pawula1967generalizations} 
for the backward process $\{x_{k+1},\dots, x_N\mid Y(t),Z(t) \}$, with transition distribution $p(y', z', t \mid y, z, t-h)$, which corresponds to the backward prior dynamics. Hence, between observation $\beta(y,z,t)$ follows the backward \ac{gfpe} 
\begin{equation*}
    \partial_t \beta(y,z,t)=-\mathcal A^\dagger \beta(y,z,t),
\end{equation*}
as derived in \cref{sec:app_derivation_hme}.

\paragraph{The Backward Distribution at Observation Time Points.}
Here, we calculate the backward distribution $\beta(y,z,t_i^-)$ right before the observation time points $\{t_i\}_{i \in 1,\dots, N}$.
We first note that
\begin{align*}
    \beta(y,z,t_i-h)&= p(x_i, \dots, x_N \mid y, z, t_i-h)\\
    &=\frac{p(x_i, \dots, x_N , y, z,t_i-h)}{p(y, z,t_i-h)}\\
    &=\frac{p(x_i\mid  x_{i+1}\dots, x_N ,  y, z,t_i-h)p(x_{i+1}, \dots, x_N , y, z,t_i-h)}{p(y, z,t_i-h)}\\
    &=p(x_i\mid x_{i+1}, \dots, x_N ,  y, z,t_i-h)p(x_{i+1}, \dots, x_N  \mid y, z,t_i-h).
\end{align*}
Calculating $h \searrow 0$, we find
\begin{equation*}
    \beta(y,z,t_i^-)=\underset{h \searrow 0}{\lim} \; \beta(y,z,t_i-h)= p(x_i \mid y) \beta(y,z,t_i).
\end{equation*}

\subsubsection{Calculation of the Smoothing Distribution}
\label{sec:app_derivation_smoothing}
We define the smoothing distribution as $\gamma(y,z,t):=p(y,z,t\mid x_{[1,N]})= C^{-1}(t)\alpha(y,z,t) \beta(y,z,t)$. 
We find the dynamics of the smoothing distribution by calculating its time derivative. For this,
we follow a proof analogous to \cite{sutter2016}. By noting that $C^{-1}(t)$ is constant almost surely \cite{pardoux1982equations},
we obtain by differentiation
\begin{align}
    &\partial_t p(y,z,t\mid x_{[1,N]})= \partial_t \gamma(y,z,t)=\partial_t \left\lbrace C^{-1}(t)\alpha(y,z,t) \beta(y,z,t)\right\rbrace \nonumber\\
    &= C^{-1}(t) \alpha(y,z,t) \partial_t \beta(y,z,t)+ C^{-1}(t) \beta(y,z,t)  \partial_t \alpha(y,z,t).
    \label{eq:ap_smoothing_ansatz}
\end{align}

The dynamics of the filtering distribution are
\begin{align*}
    \partial_t \alpha(y,z,t)=&-\sum_{i=1}^n \partial_{y_i}\left\lbrace f_i(y,z,t) \alpha(y,z,t)  \right\rbrace +\frac{1}{2} \sum_{i=1}^n \sum_{j=1}^n \partial_{y_i} \partial_{y_j}\left\lbrace D_{ij}(y,z,t) \alpha(y,z,t)\right\rbrace\\
    &{} +\sum_{z'\in \mathcal Z} \Lambda(z',z,t) \alpha(y,z',t),
\end{align*}
 where we define $\Lambda(z,z,t):=-\Lambda(z,t)$.
The dynamics of the backward distribution are given as
\begin{align*}
  \partial_t \beta(y,z,t)=&-\sum_{i=1}^n f_i(y,z,t) \partial_{y_i} \beta(y,z,t)  - \frac{1}{2} \sum_{i=1}^n \sum_{j=1} D_{ij}(y,z,t) \partial_{y_i} \partial_{y_j}\beta(y,z,t)\\
  &{} -\sum_{z'\in \mathcal Z} \Lambda(z,z',t) \beta(y,z',t).
\end{align*}

Inserting the dynamics in \cref{eq:ap_smoothing_ansatz} and using $ C^{-1}(t) \alpha(y,z,t)=\frac{\gamma(y,z,t)}{\beta(y,z,t)}$ we find
\begin{equation}
\begin{split}
 &\partial_t \gamma(y,z,t)\\
    &=\frac{\gamma(y,z,t)}{\beta(y,z,t)}\left(-\sum_{i=1}^n f_i(y,z,t) \partial_{y_i}  \beta(y,z,t) -\frac{1}{2} \sum_{i=1}^n \sum_{j=1}^n D_{ij}(y,z,t) \partial_{y_i}  \partial_{y_j} \beta(y,z,t)\right.\\
    &\quad \left.-\sum_{z'\in \mathcal Z} \Lambda(z,z',t) \beta(y,z',t)\right)\\
    &{}+\beta(y,z,t) \left(-\sum_{i=1}^n \partial_{y_i} \left\lbrace f_i(y,z,t) \frac{\gamma(y,z,t)}{\beta(y,z,t)}\right\rbrace+\frac{1}{2} \sum_{i=1}^n \sum_{j=1}^n \partial_{y_i}  \partial_{y_j} \left\lbrace D_{ij}(y,z,t) \frac{\gamma(y,z,t)}{\beta(y,z,t)}\right\rbrace\right.\\
    &\quad \left. +\sum_{z'\in \mathcal Z} \Lambda(z',z,t) \frac{\gamma(y,z',t)}{\beta(y,z',t)}\right).
    \end{split}
    \label{eq:ap_smoothing_interm1}
\end{equation}

Next we differentiate the intermediate terms using the product rule as
\begingroup
\allowdisplaybreaks
\begin{align*}
    \partial_{y_i} &\left\lbrace f_i(y,z,t) \frac{\gamma(y,z,t)}{\beta(y,z,t)}\right\rbrace\\
    =&\frac{\left(\partial_{y_i} f_i(y,z,t) \gamma(y,z,t)+f_i(y,z,t) \partial_{y_i} \gamma(y,z,t)\right)\beta(y,z,t)-f_i(y,z,t)\gamma(y,z,t) \partial_{y_i} \beta(y,z,t)}{\beta(y,z,t)^2}\\
    =&\beta(y,z,t)^{-1}\left\lbrace\partial_{y_i} f_i(y,z,t) \gamma(y,z,t)+f_i(y,z,t) \partial_{y_i} \gamma(y,z,t)\right\rbrace\\
    &{}-\beta(y,z,t)^{-2}f_i(y,z,t)\gamma(y,z,t) \partial_{y_i} \beta(y,z,t),
\end{align*}
\endgroup
and
\begingroup
\allowdisplaybreaks
\begin{align*}
    \partial_{y_i} &\partial_{y_j} \left\lbrace \frac{D_{ij}(y,z,t) \gamma(y,z,t)}{\beta(y,z,t)} \right\rbrace\\
    =&\partial_{y_i} \left\lbrace   \gamma(y,z,t) \beta(y,z,t)^{-1}\partial_{y_j} D_{ij}(y,z,t) +D_{ij}(y,z,t) \beta(y,z,t)^{-1}\partial_{y_j} \gamma(y,z,t)\right.\\
    &\qquad \left.-D_{ij}(y,z,t)\gamma(y,z,t) \beta(y,z,t)^{-2}\partial_{y_j} \beta(y,z,t)
    \right\rbrace\\
    =&\left(  \gamma(y,z,t)\partial_{y_i} \partial_{y_j} D_{ij}(y,z,t) +\partial_{y_i} D_{ij}(y,z,t) \partial_{y_j} \gamma(y,z,t) \right) \beta(y,z,t)^{-1}\\
    &-\gamma(y,z,t) \beta(y,z,t)^{-2}\partial_{y_j} D_{ij}(y,z,t)  \partial_{y_i} \beta(y,z,t) +\partial_{y_j} D_{ij}(y,z,t)\partial_{y_i}\gamma(y,z,t)\beta(y,z,t)^{-1} \\
    &+D_{ij}(y,z,t) \left(\beta(y,z,t)^{-1} \partial_{y_i} \partial_{y_j} \gamma(y,z,t) -\beta(y,z,t)^{-2} \partial_{y_j} \gamma(y,z,t) \partial_{y_i} \beta(y,z,t)\right)\\
    &-\partial_{y_i} D_{ij}(y,z,t) \gamma(y,z,t) \partial_{y_j} \beta(y,z,t) \beta(y,z,t)^{-2} \\
    &-D_{ij}(y,z,t) \left( \partial_{y_i}  \gamma(y,z,t)\partial_{y_j} \beta(y,z,t) \beta(y,z,t)^{-2}\right.\\
    &\qquad  \left. +\gamma(y,z,t)[\partial_{y_i}\partial_{y_j} \beta(y,z,t) \beta(y,z,t)^{-2}   -2 \partial_{y_j} \beta(y,z,t) \beta(y,z,t)^{-3} \partial_{y_i} \beta(y,z,t) ] \right).
\end{align*}
\endgroup
Collecting terms in $\beta(y,z,t)^{-1}$, $\beta(y,z,t)^{-2}$ and $\beta(y,z,t)^{-3}$, we find
\begin{align*}
    \partial_{y_i} &\partial_{y_j} \left\lbrace \frac{D_{ij}(y,z,t) \gamma(y,z,t)}{\beta(y,z,t)} \right\rbrace\\
    =&\beta(y,z,t)^{-1} \left\lbrace\partial_{y_i} \partial_{y_j} D_{ij}(y,z,t) \gamma(y,z,t) +\partial_{y_j} D_{ij}(y,z,t) \partial_{y_i} \gamma(y,z,t) \right. \\
    &\qquad\qquad\qquad\left.+\partial_{y_i} D_{ij}(y,z,t) \partial_{y_j} \gamma(y,z,t) +D_{ij}(y,z,t) \partial_{y_i} \partial_{y_j} \gamma(y,z,t) \right\rbrace\\
    &-\beta(y,z,t)^{-2} \left\lbrace\partial_{y_j} D_{ij}(y,z,t) \gamma(y,z,t) \partial_{y_i} \beta(y,z,t) \right.\\
    &\quad\qquad\qquad\qquad+\partial_{y_i} D_{ij}(y,z,t) \gamma(y,z,t) \partial_{y_j} \beta(y,z,t) \\
      &\quad\qquad\qquad\qquad+D_{ij}(y,z,t) \partial_{y_j} \gamma(y,z,t) \partial_{y_i} \beta(y,z,t) \\
      &\quad\qquad\qquad\qquad+D_{ij}(y,z,t) \partial_{y_i} \gamma(y,z,t) \partial_{y_j} \beta(y,z,t) \\
        &\quad\qquad\qquad\qquad \left.+D_{ij}(y,z,t) \gamma(y,z,t) \partial_{y_i}\partial_{y_j} \beta(y,z,t)\right\rbrace\\
    &+\beta(y,z,t)^{-3} \left\lbrace2 D_{ij}(y,z,t) \gamma(y,z,t) \partial_{y_i} \beta(y,z,t) \partial_{y_j} \beta(y,z,t)\right\rbrace.
\end{align*}
Using the terms in \cref{eq:ap_smoothing_interm1} we have
\begingroup
\allowdisplaybreaks
\begin{align*}
    \partial_t &\gamma(y,z,t)\\
    =&-\sum_{i=1}^n f_i(y,z,t)\gamma(y,z,t) \beta(y,z,t)^{-1} \partial_{y_i} \beta(y,z,t) \\
    &-\frac{1}{2} \sum_{i=1}^n \sum_{j=1}^n D_{ij}(y,z,t) \gamma(y,z,t) \beta(y,z,t)^{-1} \partial_{y_i} \partial_{y_j} \beta(y,z,t)\\
    &- \sum_{i=1}^n \beta(y,z,t)^{-1} \left\lbrace[\partial_{y_i} f_i(y,z,t) \gamma(y,z,t)+ f_i(y,z,t) \partial_{y_i} \gamma(y,z,t)]\beta(y,z,t)\right. \\
     &\quad\qquad\qquad\qquad \qquad\qquad\left.-f_i(y,z,t)\gamma(y,z,t) \partial_{y_i} \beta(y,z,t)\right\rbrace\\
    &+\frac{1}{2} \sum_{i=1}^n \sum_{j=1}^n \left[ \partial_{y_i} \partial_{y_j} D_{ij}(y,z,t) \gamma(y,z,t) +\partial_{y_j} D_{ij}(y,z,t) \partial_{y_i} \gamma(y,z,t) \right.\\
    &\quad\qquad\qquad\qquad +\partial_{y_i} D_{ij}(y,z,t) \partial_{y_j} \gamma(y,z,t)  +D_{ij}(y,z,t) \partial_{y_i}\partial_{y_j} \gamma(y,z,t)\\
    &\quad\qquad\qquad -\beta(y,z,t)^{-1}\left\lbrace\partial_{y_j} D_{ij}(y,z,t) \gamma(y,z,t) \partial_{y_i} \beta(y,z,t) \right. \\
    & \quad\qquad\qquad\qquad + \partial_{y_i}D_{ij}(y,z,t) \gamma(y,z,t) \partial_{y_j} \beta(y,z,t)\\
    & \quad\qquad\qquad\qquad + D_{ij}(y,z,t) \partial_{y_j} \gamma(y,z,t) \partial_{y_i} \beta(y,z,t)\\
    & \quad\qquad\qquad\qquad + D_{ij}(y,z,t) \partial_{y_i} \gamma(y,z,t) \partial_{y_j} \beta(y,z,t) \\
     & \quad\qquad\qquad\qquad\left. +D_{ij}(y,z,t) \gamma(y,z,t)\partial_{y_i}\partial_{y_j} \beta(y,z,t) \right\rbrace\\
    &\quad\qquad\qquad+\left.\beta(y,z,t)^{-2} \left\lbrace 2 D_{ij}(y,z,t) \gamma(y,z,t) \partial_{y_i} \beta(y,z,t) \partial_{y_j} \beta(y,z,t) \right\rbrace \right]\\
    &+\sum_{z'\in \mathcal Z} \left(\Lambda(z',z,t)\frac{\beta(y,z,t)}{\beta(y,z',t)} \gamma(y,z',t)-\Lambda(z,z',t)\frac{\beta(y,z',t)}{\beta(y,z,t)}\gamma(y,z,t) \right)\\
     =&-\sum_{i=1}^n \partial_{y_i} f_i(y,z,t) \gamma(y,z,t) +f_i(y,z,t) \partial_{y_i} \gamma(y,z,t)\\
     &+\frac{1}{2} \sum_{i=1}^n\sum_{j=1}^n \partial_{y_i} \partial_{y_j} D_{ij}(y,z,t) \gamma(y,z,t)+\partial_{y_j} D_{ij}(y,z,t) \partial_{y_i} \gamma(y,z,t)\\
     &\quad\qquad\qquad +\partial_{y_i}D_{ij}(y,z,t) \partial_{y_j} \gamma(y,z,t) + D_{ij}(y,z,t)\partial_{y_i}\partial_{y_j} \gamma(y,z,t)\\
     &-\frac{\beta(y,z,t)^{-1}}{2} \sum_{i=1}^n\sum_{j=1}^n \left\lbrace \partial_{y_j} D_{ij}(y,z,t) \gamma(y,z,t) \partial_{y_i} \beta(y,z,t) \right.\\
     &\qquad\qquad\qquad\qquad + \partial_{y_i}D_{ij}(y,z,t)\gamma(y,z,t) \partial_{y_j}\beta(y,z,t)\\
       &\qquad\qquad\qquad\qquad +2D_{ij}(y,z,t)\gamma(y,z,t) \partial_{y_i}\partial_{y_j} \beta(y,z,t) \\
       &\qquad\qquad\qquad\qquad +D_{ij}(y,z,t) \partial_{y_j} \gamma(y,z,t) \partial_{y_i}\beta(y,z,t)\\
       &\qquad\qquad\qquad\qquad\left.+D_{ij}(y,z,t) \partial_{y_i} \gamma(y,z,t) \partial_{y_j} \beta(y,z,t)\right\rbrace\\
     &+\beta(y,z,t)^{-2} \sum_{i=1}^n\sum_{j=1}^n  D_{ij}(y,z,t) \gamma(y,z,t) \partial_{y_i} \beta(y,z,t) \partial_{y_j} \beta(y,z,t)\\
         &+\sum_{z'\in \mathcal Z \setminus z} \left(\Lambda(z',z,t)\frac{\beta(y,z,t)}{\beta(y,z',t)} \gamma(y,z',t)-\Lambda(z,z',t)\frac{\beta(y,z',t)}{\beta(y,z,t)}\gamma(y,z,t) \right)\\
     =&-\sum_{i=1}^n \left\lbrace \partial_{y_i} f_i(y,z,t) \gamma(y,z,t) + f_i(y,z,t) \partial_{y_i} \gamma(y,z,t) \right.\\
        &\qquad\qquad+ \frac{\beta(y,z,t)^{-1}}{2} \sum_{j=1}^n \left[ \partial_{y_j} D_{ij}(y,z,t)\gamma(y,z,t) \partial_{y_i}\beta(y,z,t)\right.\\
         &\qquad\qquad\qquad\qquad\qquad\qquad+\partial_{y_i} D_{ij}(y,z,t)\gamma(y,z,t)\partial_{y_j}\beta(y,z,t)\\
        &\qquad\qquad\qquad\qquad\qquad\qquad+2D_{ij}(y,z,t)\gamma(y,z,t)\partial_{y_i}\partial_{y_j} \beta(y,z,t)\\
        &\qquad\qquad\qquad\qquad\qquad\qquad+D_{ij}(y,z,t) \partial_{y_j} \gamma(y,z,t) \partial_{y_i} \beta(y,z,t)\\
         &\qquad\qquad\qquad\qquad\qquad\qquad\left.+D_{ij}(y,z,t) \partial_{y_i} \gamma(y,z,t)\partial_{y_j} \beta(y,z,t) \right]\\
        &\qquad\qquad \left. -\beta(y,z,t)^{-2} \sum_{j=1}^n D_{ij}(y,z,t) \gamma(y,z,t) \partial_{y_i} \beta(y,z,t) \partial_{y_j} \beta(y,z,t)\right\rbrace\\
     &+\frac{1}{2} \sum_{i=1}^n \sum_{j=1}^n \partial_{y_i} \left\lbrace \partial_{y_j} D_{ij}(y,z,t) \gamma(y,z,t) +D_{ij}(y,z,t) \partial_{y_j}\gamma(y,z,t) \right\rbrace\\
         &+\sum_{z'\in \mathcal Z} \Lambda(z',z,t)\frac{\beta(y,z,t)}{\beta(y,z',t)} \gamma(y,z',t)\\
      =&-\sum_{i=1}^n \left\lbrace \partial_{y_i}\left\lbrace f_i(y,z,t) \gamma(y,z,t) \right\rbrace\right.\\
     &\qquad\qquad+\beta(y,z,t)^{-1}\sum_{j=1}^n \left[ \partial_{y_i} D_{ij}(y,z,t) \gamma(y,z,t) \partial_{y_j} \beta(y,z,t)\right.\\
     &\qquad\qquad\qquad\qquad\qquad\qquad+D_{ij}(y,z,t)\gamma(y,z,t) \partial_{y_i} \partial_{y_j}\beta(y,z,t)\\
     &\qquad\qquad\qquad\qquad\qquad\qquad+\left. D_{ij}(y,z,t) \partial_{y_i} \gamma(y,z,t) \partial_{y_j} \beta(y,z,t)\right]\\
     &\qquad\qquad \left.-\beta(y,z,t)^{-2} \sum_{j=1}^n D_{ij}(y,z,t) \gamma(y,z,t)\partial_{y_i} \beta(y,z,t) \partial_{y_j} \beta(y,z,t)\right\rbrace\\
    & +\frac{1}{2} \sum_{i=1}^n \sum_{j=1}^n \partial_{y_i} \partial_{y_j} \left\lbrace D_{ij}(y,z,t) \gamma(y,z,t) \right\rbrace +\sum_{z' \in \mathcal Z} \Lambda(z',z,t)\frac{\beta(y,z,t)}{\beta(y,z',t)} \gamma(y,z',t).
\end{align*}
\endgroup

Next, collecting terms by making use of the product rule, we have
\begingroup
\allowdisplaybreaks
\begin{align*}
    \partial_t &\gamma(y,z,t)\\
     &=-\sum_{i=1}^n \partial_{y_i} \left\lbrace f_i(y,z,t)\gamma(y,z,t) +\sum_{j=1}^n D_{ij}(y,z,t) \partial_{y_j} \beta(y,z,t) \beta(y,z,t)^{-1} \gamma(y,z,t)\right\rbrace\\
     &+\frac{1}{2} \sum_{i=1}^n \sum_{j=1}^n \partial_{y_i} \partial_{y_j} \left\lbrace D_{ij}(y,z,t) \gamma(y,z,t) \right\rbrace +\sum_{z'\in \mathcal Z} \Lambda(z',z,t)\frac{\beta(y,z,t)}{\beta(y,z',t)} \gamma(y,z',t)\\
     &=-\sum_{i=1}^n \partial_{y_i} \left\lbrace \left( f_i(y,z,t) +\sum_{j=1}^n D_{ij}(y,z,t) \partial_{y_j} \log \beta(y,z,t) \right) \gamma(y,z,t) \right\rbrace\\
      &+\frac{1}{2} \sum_{i=1}^n \sum_{j=1}^n \partial_{y_i} \partial_{y_j} \left\lbrace D_{ij}(y,z,t) \gamma(y,z,t) \right\rbrace+\sum_{z'\in \mathcal Z} \Lambda(z',z,t)\frac{\beta(y,z,t)}{\beta(y,z',t)} \gamma(y,z',t).
\end{align*}
\endgroup

Finally, we can write 
\begin{equation*}
\begin{split}
\partial_t  p(y,z,t\mid x_{[1,N]}) = &\tilde{\mathcal A} p(y,z,t\mid x_{[1,N]})\\
=&-\sum_{i=1}^n \partial_{y_i} \left\lbrace \tilde{f}_i(y,z,t)  p(y,z,t\mid x_{[1,N]}) \right\rbrace\\
&+\frac{1}{2} \sum_{i=1}^n \sum_{j=1}^n \partial_{y_i} \partial_{y_j} \left\lbrace \tilde{D}_{ij}(y,z,t)  p(y,z,t\mid x_{[1,N]}) \right\rbrace\\
&+\sum_{z'\in \mathcal Z} \tilde{\Lambda}(z',z,t)p(y,z',t\mid x_{[1,N]}),
\end{split}
\end{equation*}
with the posterior drift
\begin{align*}
\tilde{f}_i(y,z,t)=f_i(y,z,t) +\sum_{j=1}^n D_{ij}(y,z,t) \partial_{y_j} \log \beta(y,z,t),
\end{align*}
the posterior dispersion
\begin{align*}
\tilde{D}_{ij}(y,z,t)=D_{ij}(y,z,t),
\end{align*}
and the posterior rate 
\begin{align*}
 \tilde{\Lambda}(z',z,t)=\Lambda(z',z,t)\frac{\beta(y,z,t)}{\beta(y,z',t)}.
\end{align*}

\subsection{Approximate Inference}
\subsubsection{The Path-Wise \acl{kl} Divergence Between Hybrid Processes}
\label{sec:app_derivation_kl}
To derive the \ac{kl} divergence $\KL(\mathbb Q_{Y, Z}\mid\mid \mathbb P_{Y, Z})$ between two hybrid processes $\{Y^Q(t), Z^Q(t)\}_{t \geq 0}$, $\{Y^P(t), Z^P(t)\}_{t\geq 0}$
with the respective path measures $\mathbb Q_{Y, Z}, \mathbb P_{Y, Z}$,
we consider discretized versions of the continuous processes on a regular time grid,
$t \in \{0, h, 2h, ..., K\cdot h = T\}$ for some small $h$, and aim to take the limit
$h \rightarrow 0$ of the resulting expressions.

For the discretized joint paths $\{Y_k, Z_k\}_{k \in \{0, 1, ..., K\}}$,
we can explicitly write down the probability density functions; we abbreviate
$\mathbf{y} := \{y_k\}_{k\in \{0, 1, ..., K\}}$
and $\mathbf{z} := \{z_k\}_{k\in \{0, 1, ..., K\}}$ and write
\begin{align}
    \begin{split}
    q(\mathbf{y}, \mathbf{z}) &= q(y_0, z_0)\prod_{k=1}^K q(y_{k}, z_{k}\mid y_{k-1}, z_{k-1}),\\
    p(\mathbf{y}, \mathbf{z}) &= p(y_0, z_0)\prod_{k=1}^K p(y_{k}, z_{k}\mid y_{k-1}, z_{k-1}).
    \end{split}\nonumber
\end{align}
Inserting both into the \ac{kl} divergence yields
\begin{multline}\label{eq:discrete_KL}
    \KL(\mathbb Q_{Y, Z}\mid\mid \mathbb P_{Y, Z}) = \sum_{z_0, ..., z_K}\int q(y_0, z_0)
    \prod_{k=1}^K q(y_{k}, z_{k}\mid y_{k-1}, z_{k-1})\\
    \left(\ln \frac{q (y_0, z_0)}{p(z_0, y_0)} + \sum_{j=1}^K\ln \frac{q(y_{k}, z_{k}\mid y_{k-1}, z_{k-1})}
    {p(y_{k}, z_{k}\mid y_{k-1}, z_{k-1})}\right) \mathrm dy_0\cdots \mathrm dy_K \\
    = \KL(\mathbb Q^0_{Y, Z}\mid\mid \mathbb P^0_{Y, Z}) +
    \sum_{j=1}^K\sum_{z_0, ..., z_K}\int \left( q(y_0, z_0) \prod_{i=1}^j q(y_{k}, z_{k}\mid y_{k-1}, z_{k-1})\right.\\
     \left.\ln \frac{q(y_{k}, z_{k}\mid y_{k-1}, z_{k-1})}
    {p(y_{k}, z_{k}\mid y_{k-1}, z_{k-1})}\right)\mathrm dy_0\cdots \mathrm dy_K,
\end{multline}
where we introduced $\KL(\mathbb Q^0_{Y, Z}\mid\mid \mathbb P^0_{Y, Z})$
for the initial distributions:
as detailed in the main paper, we assume $q(y, z, 0) = q(z, 0) q(y, 0\mid z)$ and 
$p(y, z, 0) = p(z, 0)p(y, 0\mid z)$, which yields 
\begin{align}
    \KL(\mathbb Q^0_{Y, Z}\mid\mid \mathbb P^0_{Y, Z}) &=  \sum_z \int q(z, 0) q(y, 0\mid z) \ln 
    \frac{q(z, 0) q(y, 0\mid z)}{p(z, 0)p(y, 0\mid z)}\, \mathrm d y\nonumber\\
    &=  \sum_z \int q(z, 0) q(y, 0; z) \ln 
    \frac{q(y, 0\mid z)}{p(y, 0\mid z)}\, \mathrm dy +\sum_z q(z, 0)\ln\frac{q(z, 0)}{p(z, 0)}\nonumber\\
    &= \sum_z q(z,0) \KL(\mathbb Q^0_{Y|Z} || \mathbb P^0_{Y|Z}) + \KL(\mathbb Q^0_{Z} || \mathbb P^0_{Z}).
\end{align}
Any of the $K$ summands from the second term of \cref{eq:discrete_KL} can be simplified as
\begin{align}
    &\sum_{z_0, ..., z_K}\int \left( q(y_0, z_0) \prod_{i=1}^j q(y_{k}, z_{k}\mid y_{k-1}, z_{k-1})\right) \ln \frac{q(y_{k}, z_{k}\mid y_{k-1}, z_{k-1})}
    {p(y_{k}, z_{k}\mid y_{k-1}, z_{k-1})}\mathrm dy_0\cdots \mathrm dy_K \nonumber\\
    &=\sum_{z_{k-1},z_k}\int q(y_{k}, z_{k}\mid y_{k-1}, z_{k-1})q(y_{k-1}, z_{k-1})\ln \frac{q(y_{k}, z_{k}\mid y_{k-1}, z_{k-1})}
    {p(y_{k}, z_{k}\mid y_{k-1}, z_{k-1})} \mathrm dy_{k-1} \mathrm dy_k\nonumber\\
    &=\sum_{z_{k-1},z_k}\int q(y_{k}, z_{k}\mid y_{k-1}, z_{k-1}) q(y_{k-1}, z_{k-1})\ln  \frac{ q(y_{k}\mid y_{k-1}, z_{k-1})}{p(y_{k}\mid y_{k-1}, z_{k-1})}
    \frac{q(z_{k}\mid z_{k-1})}{p(z_{k}\mid z_{k-1})} \mathrm dy_{k-1} \mathrm dy_k\nonumber\\
    &=\sum_{z_{k-1}, z_k}\int q(y_k, z_k \mid y_{k-1}, z_{k-1})q(y_{k-1}, z_{k-1})
    \ln \frac{ q(y_{k}\mid y_{k-1}, z_k, z_{k-1})}{ p(y_{k}\mid y_{k-1}, z_k, z_{k-1})}\mathrm dy_{k-1} \mathrm dy_k\nonumber\\
    &\qquad\qquad+ \sum_{z_{k-1},z_k}  q(z_{k}, z_{k-1}) \ln\frac{ q(z_{k}\mid z_{k-1})}{ p(z_{k}\mid z_{k-1})}.\label{eq:elbo_discrete}
\end{align}
Inspecting
\begin{align*}
    q(y_k, z_k \mid y_{k-1}, z_{k-1}) &= q(y_k\mid z_k, y_{k-1}, z_{k-1})q(z_k\mid z_{k-1}),\nonumber\\
\end{align*}
and expanding $q(z_k\mid z_{k-1})$, we find
\begin{equation*}
    q(y_z, z_k\mid y_{k-1}, z_{k-1}) = 
\begin{cases}
    q(y_k\mid z_{k-1}, y_{k-1})& \text{if } z_k = z_{k-1}\\
    q(y_k\mid z_k, z_{k-1}, y_{k-1})\Lambda_{z_{k-1},z_k}h + o\left(h\right),    & \text{otherwise.}
\end{cases}
\end{equation*}
Recalling \cref{eq:characteristic_function,eq:expectation_expansion}, we furthermore have
\begin{equation*}
    q(y_k\mid z_k, z_{k-1}, y_{k-1}) = \delta(y_k - y_{k-1}) + o\left(1\right),
\end{equation*}
which can intuitively be understood as the process being continuous, thinking of $\delta(y_k - y_{k-1})$ as point-masses located at $y_k=y_{k-1}$, 
and hence, keeping only terms linear in $h$:
\begin{equation*}
    q(y_z, z_k\mid y_{k-1}, z_{k-1}) = 
\begin{cases}
    q(y_k\mid z_{k-1}, y_{k-1})& \text{if } z_k = z_{k-1}\\
    \delta(y_k - y_{k-1})\Lambda_{z_{k-1},z_k}h + o\left(h\right),    & \text{otherwise.}
\end{cases}
\end{equation*}
This yields
\begin{align*}
    &\sum_{z_0, ..., z_K}\int \left( q(y_0, z_0) \prod_{i=1}^j q(y_{k}, z_{k}\mid y_{k-1}, z_{k-1})\right) \ln \frac{q(y_{k}, z_{k}\mid y_{k-1}, z_{k-1})}
    {p(y_{k}, z_{k}\mid y_{k-1}, z_{k-1})} \mathrm dy_0\cdots \mathrm dy_K \nonumber\\
    &=\sum_{z_{k-1}}\int q(y_k\mid z_{k-1}, y_{k-1})q(y_{k-1}, z_{k-1})\ln \frac{ q(y_{k}\mid y_{k-1}, z_{k-1})}{ p(y_{k}\mid y_{k-1}, z_{k-1})} \mathrm dy_{k-1} \mathrm dy_k\nonumber\\
    &\quad + \sum_{\substack{z_{k-1}, z_k\\ z_{k-1}\neq z_k}}\int \left(\delta(y_k - y_{k-1})\Lambda_{z_{k-1},z_k}h + o\left(h\right)\right)q(y_{k-1}, z_{k-1})\nonumber\\
    &\qquad\qquad\qquad\qquad\qquad\qquad \cdot \ln \frac{ q(y_{k}\mid y_{k-1}, z_k, z_{k-1})}{ p(y_{k}\mid y_{k-1}, z_k, z_{k-1})} \mathrm dy_{k-1} \mathrm dy_k
\end{align*}
Because of $\delta(y_k-y_{k-1})$, the integrand will only contribute for $y_k = y_{k-1}$; however, 
we also have 
\begin{align*}
    \ln \frac{ q(y_{k}\mid y_{k-1}, z_k, z_{k-1})}{ p(y_{k}\mid y_{k-1}, z_k, z_{k-1})} = \ln \frac{\delta(y_k - y_{k-1}) + o(1)}{\delta(y_k - y_{k-1}) + o(1)} \overset{h \rightarrow 0}{\longrightarrow} 0.
\end{align*}

Inspecting the other log-fraction, we find
\begin{align}
    \begin{split}
    \frac{q(y_{k}\mid y_{k-1}, z_{k-1})}{p(y_{k}\mid y_{k-1}, z_{k-1})} &=
    \frac{\exp\{-\frac{1}{2h}\Vert y_{k} - y_{k-1} - g(y_{k-1}, z_{k-1}, (k-1)h)\cdot h\Vert^2_{D^{-1}}\}}
    {\exp\{-\frac{1}{2h}\Vert y_{k} - y_{k-1} - f(y_{k-1}, z_{k-1}, (k-1)h) \cdot h\Vert^2_{D^{-1}}\}}\nonumber\\
    &= \exp\left\{\frac{h}{2} \cdot \Vert g(y_{k-1}, z_{k-1}, (k-1)h) - f(y_{k-1}, z_{k-1}, (k-1)h)\Vert^2_{D^{-1}}\right\}
    \end{split}
\end{align}
and hence the ratio does not depend on the time step $k$, but only $k-1$. We hence can integrate out $z_k, y_k$.
Note that 
the normalizing prefactor $(2\pi)^\frac{k}{2}\vert D\vert^{-\frac{1}{2}}$ is the same 
for both distributions and hence cancels out; if the processes $\mathbb{Q}$ and $\mathbb{P}$
had different dispersions $D_{\mathbb{Q}}$ and $D_{\mathbb{P}}$, this cancellation would not occur and the \ac{kl} would diverge \cite{archambeauApproximateInferenceContinuousTime2011}.

Taking the limit
$K\rightarrow\infty$, $h\rightarrow 0$ with $K\cdot h = T$ yields an integral expression:
\begin{align*}
    &\int_0^T \sum_{z}\int q(y, z, t)\frac{1}{2} \Vert g(y, z, t) - f(y, z, t) \Vert_{D^{-1}}^2\,\mathrm dy \mathrm dt\nonumber\\
    &= \frac{1}{2}\int_0^T\E\left[\Vert(g(y, z, t) - f(y, z, t)\Vert_{D^{-1}}^2\right]\mathrm dt.
\end{align*}
The second term of \cref{eq:elbo_discrete} does not depend on the $Y$-process 
and hence is simply the \ac{kl} divergence between two \acp{mjp}.
Its derivation is completely analogous to the one presented above, which is why 
we omit the details and refer the interested reader to, e.g., \cite{opperVariationalInferenceMarkov2008}.
The \ac{kl} divergence is found to be
\begin{align}
    &\int_0^T \sum_{z}q_Z(z, t)\left[\sum_{z' \in \mathcal Z \setminus z} \left\lbrace \tilde{\Lambda}(z,z',t)\left(\ln \tilde{\Lambda}(z,z',t) -\ln\Lambda(z,z',t) \right)\right\rbrace-(\tilde{\Lambda}(z,t)-\Lambda(z,t))\right]\,\mathrm d t\nonumber\\
    &= \int_0^T \E\left[\sum_{z' \in \mathcal Z \setminus z} \left\lbrace \tilde{\Lambda}(z,z',t)\left(\ln \tilde{\Lambda}(z,z',t) -\ln\Lambda(z,z',t) \right)\right\rbrace-(\tilde{\Lambda}(z,t)-\Lambda(z,t))\right]
\end{align}
with the variational rates $\tilde{\Lambda}$ and the prior rates $\Lambda$ as defined 
in the main text. The full \ac{kl} divergence finally reads
\begin{multline}\label{eq:full_KL}
   \KL\left(\mathbb Q_{Y, Z} \mid\mid \mathbb P_{Y, Z}\right) = \KL(\mathbb Q^0_{Y, Z}\mid\mid \mathbb P^0_{Y, Z})+ \frac{1}{2}\int_0^T \E\left[\Vert(g(y, z, t) - f(y, z, t)\Vert_{D^{-1}}^2\right.\\
    +\left.\sum_{z' \in \mathcal Z \setminus z} \left\lbrace \tilde{\Lambda}(z,z',t)\left(\ln \tilde{\Lambda}(z,z',t) -\ln\Lambda(z,z',t) \right)\right\rbrace-(\tilde{\Lambda}(z,t)-\Lambda(z,t))\right]\,\mathrm d t.
\end{multline}

\subsubsection{Derivation of the Constrained Variational Dynamics}
\label{sec:app_derivation_approximation}
The separate dynamic constraints on the variational factors $q_Z(z, t), q_Y(y, t\mid z)$
do not in general ensure the \ac{gfpe} to be fulfilled. For meta-stable systems, however,
this is a reasonable approximation; to see this, consider the \ac{gfpe}
\begin{equation}
\begin{split}
    \partial_t q(y, z, t ) =&- \sum_{i=1}^n \partial_{y_i} \left\lbrace g_i(y,z, t) q(y, z, t )\right\rbrace \\
    &{}+  \frac{1}{2}\sum_{i=1}^n \sum_{j=1}^n
    \partial_{y_i}  \partial_{y_j} \lbrace  \nonumber     D_{ij} q(y, z, t ) \rbrace + \sum_{z^\prime \in \mathcal{Z}} \tilde{\Lambda}(z',z,t) q(y, z', t ),
    \end{split}
\end{equation}
and insert the structured mean-field approximation $q(y, z, t) = q_Z(z, t)q_Y(y, t\mid z)$:
\begin{align}
    \partial_t q(y, z, t) &= \partial_t \lbrace q_Z(z, t)q_Y(y, t\mid z)\rbrace\nonumber\\
    &= q_Y(y, t\mid z)\partial_t q_Z(z, t) + q_Z(z, t)\partial_t q_Y(y, t\mid z)\nonumber\\
    &= q_Z(z, t)\left(- \sum_{i=1}^n \partial_{y_i} \left\lbrace g_i(y,z, t) q_Y(y, t\mid z )\right\rbrace +  \frac{1}{2}\sum_{i=1}^n \sum_{j=1}^n
    \partial_{y_i}  \partial_{y_j} \lbrace  \nonumber     D_{ij} q_Y(y, t\mid z ) \rbrace\right)\nonumber\\
    &\qquad + \sum_{z' \in \mathcal Z \setminus z} \tilde{\Lambda}(z',z, t)q_Z(z',t)q_Y(y, t\mid z')-\tilde{\Lambda}(z,t)q_Z(z,t)q_Y(y, t\mid z).
\end{align}
Collecting terms, we find
\begin{align}\label{eq:ssde_constrained}
    &q_Z(z, t)\left(\partial_t q_Y(y, t\mid z)+ \sum_{i=1}^n \partial_{y_i} \left\lbrace g_i(y,z, t) q_Y(y, t\mid z )\right\rbrace -  \frac{1}{2}\sum_{i=1}^n \sum_{j=1}^n
    \partial_{y_i}  \partial_{y_j} \lbrace  \nonumber     D_{ij} q_Y(y, t\mid z ) \rbrace \right)\\
    &=  - q_Y(y, t\mid z)\partial_t q_Z(z, t) + \sum_{z' \in \mathcal Z \setminus z} \tilde{\Lambda}(z',z, t)q_Z(z',t)q_Y(y, t\mid z')-\tilde{\Lambda}(z,t)q_Z(z,t)q_Y(y, t\mid z).
\end{align}
If we are almost certain to be in state $z$ at time $t$, $q(z, t) \approx 1$, 
and the exit rate $\tilde{\Lambda}(z, t)$ from this state is small, i.e., the remain
time is large and the state is meta-stable, we have
\begin{align}
    &- q_Y(y, t\mid z)\partial_t q_Z(z, t) + \sum_{z' \in \mathcal Z \setminus z} \tilde{\Lambda}(z',z, t)q_Z(z',t)q_Y(y, t\mid z')-\tilde{\Lambda}(z,t)q_Z(z,t)q_Y(y, t\mid z)\nonumber\\
    &\approx-q_Y(y, t\mid z)\partial_z q_Z(z, t) + \tilde{\Lambda}(z, t)q_Z(z, t)q_Y(y, t\mid z) \nonumber\\
    &= q_Y(y, t\mid z)\underbrace{\left(-\partial_t q_Z(z, t) + \tilde{\Lambda}(z, t)q_Z(z, t)\right)}_{\approx 0},
\end{align}
since we know that the master equation holds (see above).
Accordingly, both sides of \cref{eq:ssde_constrained} have to vanish, that is, 
$q_Z(z, t)$ and $q_Y(y, t\mid z)$ have to individually follow the master equation and the \ac{fpe}.
The higher the uncertainty in the mode assignment $q(z, t)$, the larger the approximation error; we expect the approximation to be of high quality in regions where $z$ does not 
change rapidly.
This is acceptable, since we are genuinely interested in meta-stable systems, which by 
definition only sparingly transition between distinct, qualitatively different modes.

\subsubsection{Computing the Optimal Variational Distribution}
\label{sec:app_derivation_optimzing_objective}
We restate the ELBO $\mathcal{L}$ \cref{eq:elbo} as well as the full Lagrangian $L$ \cref{eq:Lagrangian} utilizing the shorthand notation $g - f := g(y, z, t) - f(y, z, t)$:
\begin{equation}
    \mathcal{L} = \int_0^T \ell_{\mathbb Q}(t)\,\mathrm dt,
\end{equation}
with 
\begin{align}
\begin{split}\label{eq:elbo_integrand}
    \ell_{\mathbb Q}(t) &= -\E\left[\frac{1}{2}\Vert(g - f)\Vert^2_{D^{-1}}  + \sum_{z' \in \mathcal Z \setminus z} \left\lbrace \tilde{\Lambda}(z,z',t)\ln \frac{\tilde{\Lambda}(z,z',t)}{\Lambda(z,z',t)}  \right\rbrace-(\tilde{\Lambda}(z,t)-\Lambda(z,t))\right]\\
    &{}\quad - \E\left[\ln\frac{q(y, z, 0)}{p(y, z, 0)}\right]\delta(t - 0) + \sum_{i=1}^N \E_{\mathbb Q_{Y,Z}}\left[\ln p(x_{i}\mid y_{i})\right]\delta(t - t_i),
    \end{split}
\end{align}
and 
\begin{equation}
    L = \int_0^T \ell(t)\,\mathrm dt,
\end{equation}
where
\begin{align}
    &\ell(t) = \ell_{\mathbb Q} + \sum_{z \in \mathcal{Z}}\left[\lambda^\top(z, t)\left( \dot{\mu}(z, t)  - \right. 
        \left(A(z, t)\mu(z, t) + b(z, t) \right)\right)\label{eq:Lagrangian_integrand}\\
        &{}+\tr\left\{\Psi^\top(z, t) \left(\dot{\Sigma}(z, t)
        -\left(A(z, t)\Sigma(z, t) + \Sigma(z, t)A^\top(z, t) + D\right)\right)\right\}\nonumber\\
        &{}+ \left.\nu(z,t) \left(\dot{q}_Z(z,t) - \sum_{z' \in \mathcal Z} \tilde{\Lambda}_{z'z}(t)q_Z(z',t) \right)\right]\nonumber.
\end{align}

\paragraph{Stationarity with respect to the Variational MJP}
The \ac{el} equation for $q_Z$ reads
\begin{equation*}
    \frac{\mathrm d}{\mathrm d t}\frac{\partial\ell}{\partial \dot{q}_Z} = \frac{\partial \ell}{\partial q_Z}.
\end{equation*}
With \cref{eq:Lagrangian_integrand}, we therefore find the components
\begin{equation*}
    \frac{\mathrm d}{\mathrm dt}\frac{\partial \ell}{\partial \dot{q}_Z(z, t)} = \frac{\mathrm d}{\mathrm dt}\nu(z, t)
\end{equation*}
and
\begin{align*}
    \frac{\partial \ell}{\partial q_Z(z, t)} &= \frac{\partial\ell_{\mathbb Q}}{\partial q_Z(z, t)} - \sum_{z'\in\mathcal Z \setminus z}\tilde{\Lambda}(z, z', t)\nu(z', t) + \tilde{\Lambda}(z, t)\nu(z, t).
\end{align*}
Hence, the \ac{el} equation
yields
\begin{equation}
     \frac{\mathrm d}{\mathrm d t} \nu(z, t) = \frac{\partial\ell_{\mathbb Q}}{\partial q_Z(z, t)} - \sum_{z'\in\mathcal Z \setminus z}\tilde{\Lambda}(z, z', t)\nu(z', t) + \tilde{\Lambda}(z, t)\nu(z, t).
\end{equation}

Using the law of total expectation $$\E\left[\phi(Y(t),Z(t),t)\right]=\sum_z q_Z(z,t) \E\left[\phi(Y(t),Z(t),t)\mid Z(t)=z\right],$$ we calculate
the gradient of \cref{eq:elbo_integrand} with respect to $q_Z(z, t)$ as
\begin{equation*}
\begin{split}
    \partial_{q_Z(z, t)}\ell_{\mathbb Q} &= -\E\left[\Vert g-f\Vert^2_{D^{-1}} \vert z\right]\\
    &{}\quad - \sum_{z'\in \mathcal Z \setminus z}\left[\tilde{\Lambda}_{zz'}(t)\left(\ln\frac{\tilde{\Lambda}_{zz'}(t)}{\Lambda_{zz'}} - 1 + \nu(z', t) - \nu(z, t)\right) + \Lambda_{zz'}\right]\\
    &{}\quad + \sum_i^N \E\left[\ln p(x_{i}\mid y_{i})\right\vert z]\delta(t - t_i).
    \end{split}
\end{equation*}
For linear prior drift functions $f(y, z, t) = A_p(z, t)y + b_p(z, t)$,  we can calculate $\E[\Vert g - f \Vert^2_{D^{-1}}\vert z]$ explicitly.
We use $\bar{A}(z, t) := A(z, t) - A_p(z, t)$ and $\bar{b}(z, t) := b(z, t) - b_p(z, t)$ and obtain
\begin{equation*}
\begin{split}
       \E\left[\Vert g - f \Vert^2_{D^{-1}}\vert z\right]&= \tr\{\bar{A}(z, t)^\top D^{-1} \bar{A}(z, t)\Sigma(z, t)\} \\
    &\qquad +   (\bar{A}(z, t)\mu(z, t) + \bar{b}(z, t))^\top D^{-1}(\bar{A}(z, t)\mu(z, t) + \bar{b}(z, t)). 
\end{split}
\end{equation*}
Therefore, the gradient $ \partial_{q_Z(z, t)}\ell_{\mathbb Q}$ reads
\begin{equation*}
    \begin{split}
    \partial_{q_Z(z, t)}\ell_{\mathbb Q} &= -\tr\{\bar{A}(z, t)^\top D^{-1} \bar{A}(z, t)\Sigma(z, t)\} \\
    &\qquad -   (\bar{A}(z, t)\mu(z, t) + \bar{b}(z, t))^\top D^{-1}(\bar{A}(z, t)\mu(z, t) + \bar{b}(z, t))\\
    &{}\quad - \sum_{z'\in \mathcal Z \setminus z}\left[\tilde{\Lambda}_{zz'}(t)\left(\ln\frac{\tilde{\Lambda}_{zz'}(t)}{\Lambda_{zz'}} - 1 + \nu(z', t) - \nu(z, t)\right) + \Lambda_{zz'}\right]\\
    &{}\quad + \sum_{i=1}^N \E\left[\ln p(x_{i}\mid y_{i})\right\vert z]\delta(t - t_i).
    \end{split}
\end{equation*}

Due to the delta-contributions $\delta(t - t_i)$, the evolution equation for the Lagrange multiplier $\nu(z, t)$
\cref{eq:MJP_lagrange_equation} is an impulsive differential equation
\cite{samoilenko1995impulsive} which can be solved piece-wise by integrating 
the \ac{ode} between the discontinuities (starting at $\nu(z, T) = 0$) and
applying reset conditions at the integration boundaries,
similar to exact posterior inference (c.f. \cref{sec:app_derivation_smoothing}):
\begin{align}
    \nu(z, t_i) = \E\left[\ln p(x_{i}\mid y_{i})\right\vert z] + \nu(z, t_i^-), 
\end{align}
with  $\nu(z, t_i^-):= \lim_{h \searrow 0} \nu(z, t_i - h)$.

For Gaussian observation noise, we have
\begin{multline}
    \E\left[\ln p(x_i\mid y_i)\vert z\right] = -\frac{1}{2}\left\{n\ln(2\pi) + \ln\vert\Sigma_{\mathrm{obs}}\vert\right.\\
     \left.+(x_i - \mu(z, t_i))^\top \Sigma^{-1}_{\mathrm{obs}}(x_i - \mu(z, t_i)) + \tr\{\Sigma^{-1}_{\mathrm{obs}}\Sigma(z, t_i)\}\right\}.
\end{multline}

\paragraph{Stationarity with respect to the Variational \acp{gp}}

In the same manner as for the variational \ac{mjp}, we straightforwardly arrive at
\begin{equation}
    \begin{split}
        \frac{\mathrm d}{\mathrm d t}\lambda(z, t)&=\partial_{\mu(z,t)}  \ell_{\mathbb Q}- A^\top(z, t)\lambda(z, t),\\
        \frac{\mathrm d}{\mathrm d t}\Psi(z, t)&= \partial_{\Sigma(z, t)}  \ell_{\mathbb Q} - A^\top(z, t)\Psi(z, t) 
        - \Psi(z, t)A(z, t).
    \end{split}
\end{equation}
We find the gradients as
\begin{equation*}
    \begin{split}
         \partial_{\mu(z,t)}  \ell_{\mathbb Q} &= -\partial_{\mu(z, t)}\E\left[\Vert g - f \Vert^2_{D^{-1}}\right] + \sum_{i=1}^N \partial_{\mu(z, t)}\E[\ln p(x_i\mid y_i)]\delta(t - t_i),\\
         \partial_{\Sigma(z,t)} \ell_{\mathbb Q} &= -\partial_{\Sigma(z, t)}\E\left[\Vert g - f \Vert^2_{D^{-1}}\right] + \sum_{i=1}^N \partial_{\Sigma(z, t)}\E[\ln p(x_i\mid y_i)]\delta(t - t_i).
    \end{split}
\end{equation*}
We compute $\E\left[\Vert g - f \Vert^2_{D^{-1}}\right]$ for linear prior models as
\begin{equation}
\label{eq:SDE_contribution}
\begin{split}
       \E\left[\Vert g - f \Vert^2_{D^{-1}}\right] &= \E\left[(\bar{A}(z, t) y + \bar{b}(z, t))^\top D^{-1}(\bar{A}(z, t)y + \bar{b}(z, t))\right]\nonumber\\
    &= \sum_z q(z, t)\left\{ \tr\{\bar{A}(z, t)^\top D^{-1} \bar{A}(z, t)\Sigma(z, t)\} \right.\\
    &\qquad + \left. (\bar{A}(z, t)\mu(z, t) + \bar{b}(z, t))^\top D^{-1}(\bar{A}(z, t)\mu(z, t) + \bar{b}(z, t)) \right\}.
\end{split}
\end{equation}
Using the Gaussian observation likelihoods we arrive at
\begin{equation*}
   \begin{split}
       \partial_{\mu(z,t)}  \ell_{\mathbb Q} &=-q_Z(z, t) \left(\bar{A}(z, t)^\top D^{-1}\bar{A}(z, t)\mu(z, t) + \bar{A}(z, t)^\top D^{-1}\bar{b}(z, t)\right)\\
    &{}\quad +\sum_{i=1}^N q_Z(z, t_i)\Sigma^{-1}_{\mathrm{obs}}(x_i - \mu(z, t_i))\delta(t - t_i),\\
    \partial_{\Sigma(z,t)} \ell_{\mathbb Q} &=-\frac{1}{2}q_Z(z, t)\bar{A}^\top(z, t)D^{-1}\bar{A}(z, t) - \sum_{i=1}^N q(z, t_i)\frac{1}{2}\Sigma^{-1}_{\mathrm{obs}}\delta(t - t_i).
   \end{split} 
\end{equation*}
The solutions to these impulsive differential equations are found as for the 
Lagrange multiplier $\nu(z, t)$, with reset conditions
\begin{equation*}
\begin{split}
    \lambda(z, t_i)&=q_Z(z, t_i)\Sigma^{-1}_{\mathrm{obs}}(x_i - \mu(z, t_i)) + \lambda(z, t_i^-) \\
    \Psi(z, t_i) &= -q_Z(z, t_i)\frac{1}{2}\Sigma^{-1}_{\mathrm{obs}} + \Psi(z, t_i^-),
    \end{split}
\end{equation*}
with $\lambda(z, t_i^-):= \lim_{h\searrow 0}\lambda(z, t_i - h)$ and  $ \Psi(z, t_i^-):= \lim_{h\searrow 0}\Psi(z, t_i - h)$.

\subsubsection{The Optimal Variational Parameters}
\label{sec:app_control_gradients}
To optimize the variational parameters, we employ a heuristic back-tracking line search
algorithm, as is standard in the field
\cite{bertsekas1997nonlinear}: we choose as step size $\kappa_i=\gamma^i$, where the exponential decay factor is chosen as $\gamma=0.5$. We therefore update the
current parameter $u(t) \in \{A, b, \tilde{\Lambda}\}$ as 
\[
u_{\mathrm{new}}(t) = u(t) + \kappa_i\cdot \partial_{u(t)}\ell.
\]
If  $\mathcal{L}[u_{\mathrm{new}}(t)] \geq \mathcal{L}[u(t)]$, we accept the update.
Otherwise, we iterate and re-compute using the new step size $\kappa_{i+1}$. 

\paragraph{Gradients for $A(z, t), b(z, t), \tilde{\Lambda}$}
We provide here the explicit gradients with respect to 
the variational parameters. We have
\begin{align}
    \partial_{A(z, t)} L &= - \frac{1}{2}\partial_{A(z, t)}\E\left[\Vert g - f \Vert^2_{D^{-1}}\right] - \lambda^\top(z, t) \mu(z, t) - 2\Psi(z, t)\Sigma(z, t)\nonumber\\
    \begin{split}
         &= -q_Z(z, t)D^{-1}\left(\bar{A}(z, t)(\mu(z, t)\mu^\top(z, t) + \Sigma(z, t)) + \bar{b}(z, t)\mu^\top(z, t)\right) \\
         &{}\quad- \lambda^\top(z, t) \mu(z, t) - 2\Psi(z, t)\Sigma(z, t).
    \end{split}
\end{align}

Similarly, we find
\begin{align}
    \partial_{b(z, t)} L &= -\partial_{b_q(z, t)} \frac{1}{2}\E\left[\Vert g - f \Vert^2_{D^{-1}}\right] - \lambda(z, t)\nonumber\\
    &= -q_Z(z, t) D^{-1}\left(\bar{A}(z, t)\mu(z, t) + \bar{b}(z, t)\right) - \lambda(z, t).
\end{align}

Finally,
\begin{align}
    \partial_{\tilde{\Lambda}_{zz'}(t)}L &= -\partial_{\tilde{\Lambda}_{zz'}(t)} \left[\sum_{z' \in \mathcal Z \setminus z} \left\lbrace \tilde{\Lambda}(z,z',t)\ln \frac{\tilde{\Lambda}(z,z',t)}{\Lambda(z,z',t)}  \right\rbrace-(\tilde{\Lambda}(z,t)-\Lambda(z,t))\right]\nonumber\\
    &{}\quad + \nu(z, t)q_Z(z, t) - \nu(z', t)q_Z(z, t)\nonumber\\
    &= q_Z(z, t)\left(-\ln\frac{\tilde{\Lambda}_{zz'}(t)}{\Lambda_{zz'}} + \nu(z, t) - \nu(z', t)\right).
\end{align}

\paragraph{Variational Initial Conditions}
The gradients with respect to the initial conditions result from Pontryagin's maximum principle
\cite{liberzonCalculusVariationsOptimal2012,bertsekas1997nonlinear} as
\begin{equation}
\begin{split}
    \partial_{\mu(z,0)} L &= \partial_{\mu(z,0)} \KL(\mathbb{Q}^0_{Y,Z} || \mathbb{P}^0_{Y,Z}) + \lambda(z,0) = 0,\\
    \partial_{\Sigma(z,0)}L &= \partial_{\Sigma(z,0)} \KL(\mathbb{Q}^0_{Y,Z} || \mathbb{P}^0_{Y,Z})  + \Psi(z,0) = 0,\\
    \partial_{q_Z(z,0)}L &= \partial_{q_Z(z,0)} \KL(\mathbb{Q}^0_{Y,Z} || \mathbb{P}^0_{Y,Z})  + \nu(z,0) = 0.
\end{split}
\end{equation}

While in principle, one could use these expressions to find closed-form solutions for the 
initial parameters, this is not possible in practice for $\Sigma(z, 0)$ and $q_Z(z, 0)$. Also,
resetting the parameters may cause numerical instabilities in the forward-backward sweeping algorithm
for the constraints and the Lagrange multipliers, cf. \cref{sec:inference}. We hence utilize the 
same gradient ascent update scheme as above.

We assume a Gaussian prior initial distribution, i.e. $p(y,0 \mid z)=\mathcal N(y \mid \mu_p^0(z), \Sigma_p^0(z))$, 
and for the presented variational ansatz we have an initial variational distribution $q(y,0 \mid z)=\mathcal N(y \mid \mu(z,0), \Sigma(z,0))$, which is also Gaussian.
This yields
\begin{equation*}
\begin{split}
    \KL(\mathbb{Q}^0_{Y|Z} \mid\mid \mathbb{P}^0_{Y|Z}) &= \KL(\mathcal N(y \mid \mu(z,0), \Sigma(z,0)) || \mathcal N(y \mid \mu_p^0(z), \Sigma_p^0(z))\\
    &= \frac{1}{2} \left\lbrace \ln \frac{\vert \Sigma_p^0(z) \vert}{\vert \Sigma(z,0) \vert} +\tr\left(\Sigma_p^0(z)^{-1} \Sigma(z,0)\right) \right. \\
    &{}\quad \left. + \left( \mu_p^0(z)-\mu(z,0) \right) \Sigma_p^0(z)^{-1} \left( \mu_p^0(z)-\mu(z,0) \right)^\top  + n\right\rbrace.
\end{split}
\end{equation*}

We readily compute
\begin{equation}
\begin{split}
    \partial_{\mu(z,0)} \KL(\mathbb{Q}^0_{Y,Z} \Vert \mathbb{P}^0_{Y,Z}) &=q_Z(z,0) \Sigma_p^0(z)^{-1} (\mu(z,0)-\mu_p^0(z)),\\
    \partial_{q_Z(z,0)} \KL(\mathbb{Q}^0_{Y,Z} \Vert \mathbb{P}^0_{Y,Z}) &= \partial_{q_Z(z,0)} \left\lbrace \KL(\mathbb Q^0_{Z} \Vert \mathbb P^0_{Z}) \right\rbrace  + \KL(\mathbb Q^0_{Y|Z} \Vert \mathbb P^0_{Y|Z}).
    \end{split}
\end{equation}

For the covariance matrix $\Sigma(z, 0)$ and the initial distribution $q_Z(z, 0)$, 
we require additional constraints. The covariance $\Sigma(z, 0)$ needs to be positive
semi-definite, which can be enforced by a reparameterization as $\Sigma(z,0)=CC^\top$.
We calculate the gradient with respect to the objective 
\begin{equation*}
    \mathcal L_{\mathrm{aug}}(C)= q(z,0)  \left\lbrace \KL(\mathcal N(y \mid \mu(z,0), CC^\top) || \mathcal N(y \mid \mu_p^0(z), \Sigma_p^0(z)) \right\rbrace + \tr \left( \Psi(z,0)^\top C C^\top \right)
\end{equation*}
and the PyTorch package for automatic differentiation and optimization \cite{NEURIPS2019_9015}.

The initial distribution $q_Z(z, t)$ needs to fulfil $\sum_z q_Z(z, 0) = 1$, so we 
optimize an augmented cost function
\begin{align*}
    \mathcal L_{\mathrm{aug}}(\mathbb{Q}^0_{Z},\xi)&=\KL(\mathbb{Q}^0_{Z} || \mathbb{P}^0_{Z}) +\sum_z q_Z(z,0) \KL(\mathbb Q^0_{Y|Z} \Vert \mathbb P^0_{Y|Z})  \\
    &{}+ \sum_z \nu(z,0) q_Z(z,0) + \xi (1-\sum_z q_Z(z,0)) ,
\end{align*}
where $\xi(z)$ are Lagrange multipliers. We can again eliminate the constraints by enforcing a
reparameterization \cite{boyd2004convex} as $q_Z(z,0)=q_z$ for $z\in \{1,\dots,k -1\}$, with
$k=\vert \mathcal Z\vert$ and $q_Z(k,0)=1-\sum_{z=1}^{k-1} q_z$, which yields the
unconstrained problem
\begin{equation*}
\begin{split}
    \mathcal L_{\mathrm{aug}}(q_1,\dots,q_{k-1})= &\sum_{z=1}^{k-1} q_z \ln \frac{q_z}{p(z,0)} + (1-\sum_{z=1}^{k-1} q_z)\ln \frac{1-\sum_{z=1}^{k-1} q_z}{p(k,0)}\\
    &{}+ \sum_{z=1}^{k-1} q_z \KL(\mathbb Q^0_{Y|Z=z} || \mathbb P^0_{Y|Z=z}) +  (1- \sum_{z=1}^{k-1} q_z) \KL(\mathbb Q^0_{Y|Z=k} || \mathbb P^0_{Y|Z=k})\\
    &{}+ \sum_{z=1}^{k-1} \nu(z,0) q_z +  \nu(k,0)(1- \sum_{z=1}^{k-1} q_z).
\end{split}
\end{equation*}

We find
\begin{equation}
\begin{split}
    \partial_{q_z} \mathcal L_{\mathrm{aug}} &=\ln \frac{q_z}{p(z,0)} + 1 - \ln \frac{1-\sum_{z=1}^{k-1} q_z}{p(k,0)} -1 \\
    &{}\quad + \KL(\mathbb Q^0_{Y|Z=z} || \mathbb P^0_{Y|Z=z})-\KL(\mathbb Q^0_{Y|Z=k} || \mathbb P^0_{Y|Z=k}) + \nu(z,0)-\nu(k,0)\\
    &=\ln  \frac{q_z p(k,0)}{(1-\sum_{z=1}^{k-1} q_z)p(z,0)}    \\
    &\quad + \KL(\mathbb Q^0_{Y|Z=z} || \mathbb P^0_{Y|Z=z})-\KL(\mathbb Q^0_{Y|Z=k} || \mathbb P^0_{Y|Z=k}) + \nu(z,0)-\nu(k,0).\\
    \end{split} 
\end{equation}

\subsubsection{Optimizing the Prior Parameters}
\label{sec:app_parameter_gradients}

The parameters of the original process $p$, see \cref{tab:parameters},
can be learned straightforwardly
by optimizing the full Lagrangian \cref{eq:Lagrangian} with respect to them.
\begin{table}
    \centering
    \begin{tabular}[b]{c|c}
        $A_p$ & prior slope\\
        $b_p$ & prior intercept\\
        $\Lambda_{zz'}$ & prior transitions rates\\
        $\Sigma_{\mathrm{obs}}$ & observation covariance\\
        $D$ & Dispersion matrix\\ 
        $\mu_p(z, 0), \Sigma_p(z, 0), p(z, 0)$ & Prior initial conditions\\
        \hline
    \end{tabular}
    \caption{Model parameters learned via \ac{vem}}
    \label{tab:parameters}
\end{table}

\paragraph{Prior \ac{mjp} transition rates}
With the usual shorthand $\Lambda_{zz'}(t) = \Lambda(z, z', t)$, we compute the
prior transition rates (which we in all cases assume to be time-homogeneous, $\Lambda_{zz'}(t) = \Lambda_{zz'}$):
\begin{align}
    \frac{\partial L}{\partial \Lambda_{ij}} &= - \frac{\partial }{\partial \Lambda_{ij}} \int_0^T \sum_z q_Z(z, t)\sum_{z' \in \mathcal Z \setminus z} \left\lbrace \tilde{\Lambda}_{zz'}(t)\ln \frac{\tilde{\Lambda}_{zz'}(t)}{\Lambda_{zz'}}  \right\rbrace-(\tilde{\Lambda}(z,t)-\Lambda(z))\mathrm d t\\
    &= \frac{\partial }{\partial \Lambda_{ij}} \int_0^T \sum_z q_Z(z, t)\left[\sum_{z' \in \mathcal Z \setminus z}\left\{\tilde{\Lambda}_{zz'}(t)\ln \Lambda_{zz'} - \Lambda_{zz''} \right\}\right]\mathrm d t\\
    & = \frac{1}{\Lambda_{ij}}\int_0^T q_Z(i, t)\tilde{\Lambda}_{ij}(t) \mathrm d t - \int_0^T q_Z(i, t)\mathrm d t .
\end{align}

Setting this to zero yields
\begin{equation}
    \Lambda_{ij} = \frac{\int_0^T q_Z(i, t)\tilde{\Lambda}_{ij}(t) \mathrm d t}{\int_0^T q_Z(i, t) \mathrm d t}.
\end{equation}

\paragraph{Observation covariance}
To determine the observation covariance, we compute
\begin{align}
    \frac{\partial L}{\partial \Sigma_{\mathrm{obs}}^{-1}} &= -\frac{\partial }{\partial \Sigma_{\mathrm{obs}}^{-1}}\E\left[\sum_i \ln p(x_i\mid y_i)\right]\\
    &=-\frac{N}{2}\Sigma_{\mathrm{obs}} + \sum_{i=1}^N\frac{1}{2}\sum_{z\in\mathcal Z}q_Z(z, t_i)\left[(x_i - \mu(z, t_i))(x_i - \mu(z, t_i))^\top + \Sigma(z, t_i)\right],
\end{align}
yielding
\begin{equation}
   \Sigma_{\mathrm{obs}} = \frac{1}{N}\sum_{i=1}^N\sum_{z\in\mathcal Z}q_Z(z, t_i)\left[(x_i - \mu(z, t_i))(x_i - \mu(z, t_i))^\top + \Sigma(z, t_i)\right].
\end{equation}

\paragraph{Dispersion} 
We update the dispersion in the same way as the variational parameters (c.f. \cref{sec:app_control_gradients}) and provide the gradient with respect to the dispersion $D$; note that
the more general mode-dependent dispersion $D(z)$ are found in the same way by omitting the summation
over $z$.
\begin{align}
    \partial_{D} L &= \partial_{D} \frac{1}{2}\int_0^T \E\left[\Vert g - f \Vert^2_{D^{-1}}\right]\mathrm dt + \partial_{D}\sum_z\int_0^T\tr\{\Psi^\top (z, t)D\}\mathrm dt\nonumber\\
    &= \frac{1}{2}\int_0^T \partial_{D}\E\left[\Vert g - f \Vert^2_{D^{-1}}\right]\mathrm dt+ \int_0^T\sum_{z\in\mathcal Z}\Psi (z, t)\mathrm dt\nonumber\\
    &= \frac{1}{2}-D^{-\top}\left(\int_0^T \sum_{z\in\mathcal{Z}}q_Z(z, t)\E[(\bar{A}(z, t)y + \bar{b})(\bar{A}(z, t)y + \bar{b})^\top\vert z]\right)D^{-\top}\\
    &{}\qquad+ \int_0^T\sum_{z\in\mathcal{Z}}\Psi (z, t)\mathrm dt.\nonumber
\end{align}

Note that the prior initial conditions $\mu_p^0(z), \Sigma_p^0(z), p(z, 0)$  trivially 
minimize their \ac{kl} divergence to the variational initial conditions by equality.

\paragraph{Prior drift parameters}
Also the prior parameters $A_p, b_p$ are learned utilizing \ac{vem}, see \cref{sec:app_control_gradients}.
The gradients are found as
\begin{align}
    \partial_{A_p(z)}L &= \partial_{A_p(z)}\frac{1}{2}\int_0^T \E\left[\Vert g - f \Vert^2_{D^{-1}}\right]\mathrm d t\\
    & = \frac{1}{2}\int_0^Tq(z, t)\partial_{A_p(z)}\left\{\tr\{\bar{A}(z, t)^\top D^{-1} \bar{A}(z, t)\Sigma(z, t)\} \right.\nonumber\\
    & = - \int_0^T q_Z(z, t)\left(D^{-1}\bar{A}(z, t)\left(\Sigma(z, t) + \mu(z, t)\mu^\top(z, t)\right) + D^{-1}\bar{b}(z, t)\mu^\top(z, t)\right)\mathrm d t,\nonumber\\
    \partial_{b_p(z)}L &= \partial_{b_p(z)}\frac{1}{2}\int_0^T \E\left[\Vert g - f \Vert^2_{D^{-1}}\right]\mathrm d t\\
    &=  \int_0^T q_Z(z, t)D^{-1}(\bar{A}(z, t)\mu(z, t) + \bar{b}(z, t))\nonumber
\end{align}

\section{Experiments}
\subsection{Model Validation on Ground-Truth Data}
\label{sec:app_exp_synSSDE}
A comprehensive overview over the ground-truth and learned parameters is given in \cref{tab:synSSDE_parameters}. We model the dispersion as constant, $D(z, t) = D$ and the
underlying prior \ac{mjp} as time-homogeneous, $\Lambda(z, z', t) = \Lambda(z, z')$.
The prior drift function reads $f(y, z, t) = \alpha_z(\beta_z - y)$. We will use this
parameterization in the following; to convert between this and the hitherto used 
$f(y, z, t) = A_p(z)y + b_p(z)$, use
\begin{align}
\begin{split}
    A_p(z) &= -\alpha_z\\
    b_p(z) &= \alpha_z\beta_z.
\end{split}
\end{align}

We draw the observation times from a Poisson point process with intensity 
$\frac{1}{\lambda} = 0.35$, meaning that the average inter-observation interval is $0.35$.

We initialize our model empirically by running a k-means algorithm with $k=2$ \cite{barber2012} 
on the observed data. Note that we utilize this procedure for all experiments.
The initial prior means and covariances, $\mu_p(z, 0)$ and $\Sigma_p(z, 0)$,
are then set as the cluster means and intra-cluster covariances. The prior intercept $b_p(z)$ is 
set in the same way. The initial observation covariance
$\Sigma_{\mathrm{obs}}$ as well as the dispersion $D$ are both set as the average of the intra-cluster covariances.
We can not easily initialize the prior rates $\Lambda$ and the prior slope $A_p(z)$ empirically. We set
\begin{equation*}
\Lambda = 
\begin{pmatrix}
-1 & 1\\
1 & -1
\end{pmatrix}
\end{equation*}
and $A_p(z) = \{-1, -1\}$. The initial prior $p(z, 0)$ is initialized uniformly. The corresponding variational
quantities such as $A(z, t)$, are initialized as constant functions on the initial value, e.g.,
\begin{equation*}
    A(z, t) = A_p(z)\,\forall t.
\end{equation*}

\begin{figure}
    \centering
    \includegraphics[width=.6\columnwidth]{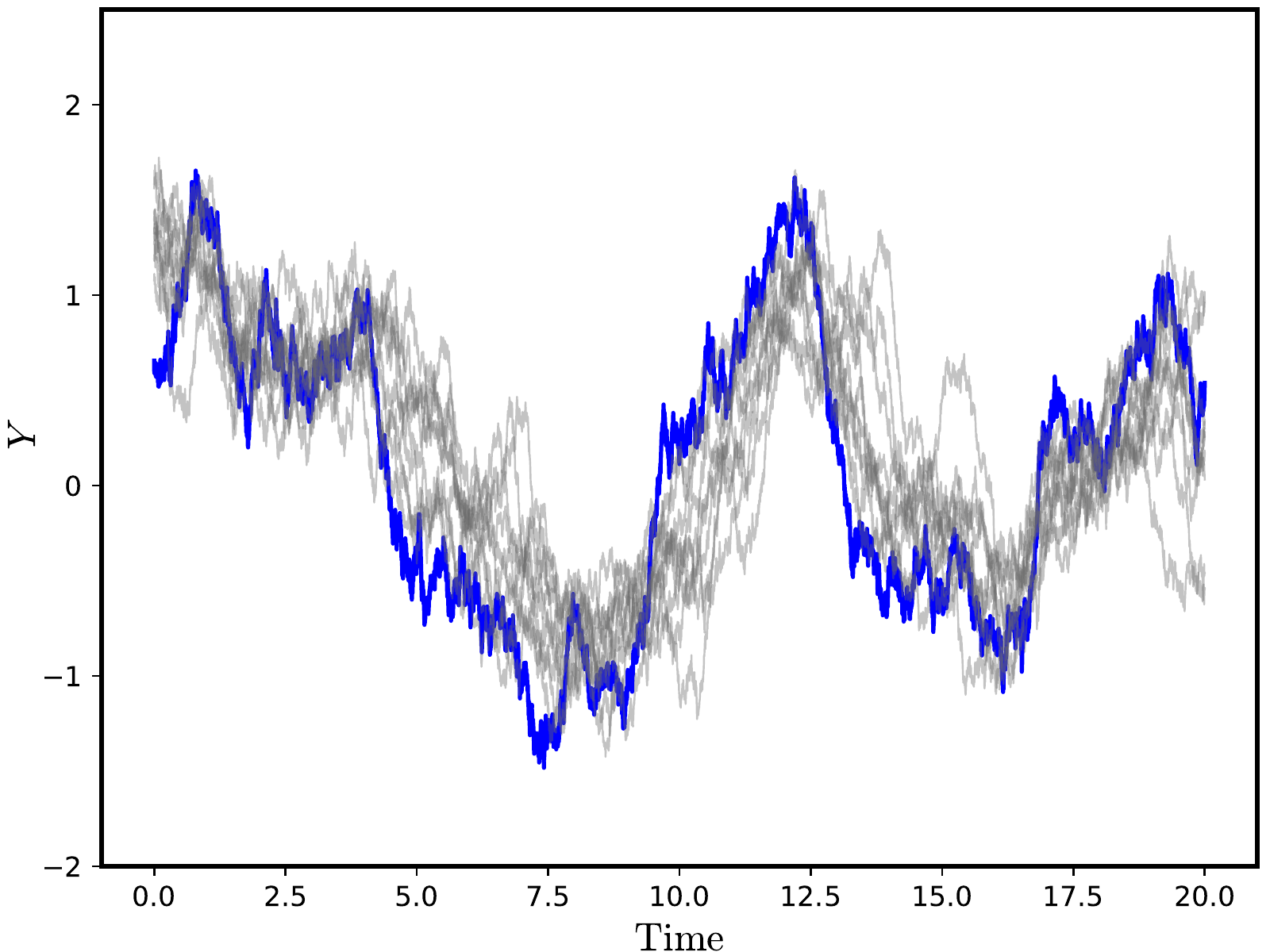}
    \caption{1D, two-mode hybrid process: posterior samples from the variational distribution (gray) and the latent ground-truth trajectory (blue).}
    \label{fig:app_fig1}
\end{figure}

We generate samples from the variational posterior to demonstrate the quality of the latent trajectory
reconstruction, see \cref{fig:app_fig1}. To sample from the posterior \ac{mjp} with time-inhomogeneous rates
$\tilde{\Lambda}$, we utilize the thinning algorithm \cite{lewis1979simulation}.

\begin{table}
    \centering
    \begin{tabular}[b]{c|c|c}
        Parameter & Ground truth value & Learned value \\
        \hline
        $\alpha_z$ & $[1.5, 1.5]$ & $[0.91, 1.28]$\\
        $\beta_z$ & $[-1, +1]$ & $[-0.72, 0.41]$ \\
        $\Lambda$ & $\begin{pmatrix} -0.2 & 0.2 \\0.2 & -0.2 \end{pmatrix}$ & $\begin{pmatrix} -0.64 & 0.64 \\0.63 & -0.63 \end{pmatrix}$\\
        $\Sigma_{\mathrm{obs}}$ & 0.1 & 0.21 \\
        $D$ & 0.25 & 0.17 \\
        $\mu_p(z, 0)$ & $[-1, +1]$ & $[-0.50, 1.39]$\\
        $\Sigma_p(z, 0)$ & $[0.2, 0.2]$ & $[0.27, 0.03]$ \\
        $p(z, 0)$ & $[0, 1]$ & $[0, 1]$\\
        \hline
    \end{tabular}
    \caption{Ground truth and learned parameters of the 1D, two-mode hybrid process.}
    \label{tab:synSSDE_parameters}
\end{table}

\subsection{Diffusions in Multi-Well Potentials}
\label{sec:app_exp_diffusions}

The one-dimensional 4-well potential is given as 
\begin{equation}
    V(y) = 4\left(y^8 + 3e^{-80 y^2} + 2.5e^{-80(y-0.5)^2)}+ 2.5e^{-80(y+0.5)^2)}\right),
\end{equation}
the two-dimensional 3-well potential reads
\begin{multline}
    V(y_1, y_2) = 3 e^{-y_1^2 - (y_2 - \frac{1}{3})^2} - 3e^{-y_1^2 - (y_2 - \frac{5}{3})^2}
    -5 e^{-(y_1 - 1)^2 - y_2^2}\\ - 5e^{-(y_1+1)^2 - y_2^2} + 0.2y_1^4 + 0.2\left(y_2 - \frac{1}{3}\right)^4.
\end{multline}

In the 1D example, we fix the observation covariance as $\Sigma_{\mathrm{obs}} = 0.0225$. We 
provide the list of all learned parameters in \cref{tab:1D_4well}.
In the 2D example, we fix $\Sigma_{\mathrm{obs}} = \begin{pmatrix} 0.2 & 0\\ 0 & 0.2
\end{pmatrix}$; the exhaustive list of parameters is given in \cref{tab:2D_3well}.
The initialization is done as in \cref{sec:app_exp_synSSDE}.

\begin{table}
    \centering
    \begin{tabular}[b]{c|c}
        Parameter & Learned value \\
        \hline
        $\alpha_z$ & $[1.11, 0.94, 0.99, 0.99]$ \\
        $\beta_z$ & $[0.79, -0.27, 0.25, -0.69]$\\
        $\Lambda$ & $\begin{pmatrix} -0.69 & 0.14 & 0.48 & 0.07 \\ 0.07 & -1.39 & 0.45 & 0.87\\
        1.11 & 0.87 & -2.33 & 0.35 \\
        0.04 & 1.44& 0.50 & -1.98\end{pmatrix}$\\
        $D$ & $[0.015, 0.001, 0.003, 0.01]$ \\
        $\mu_p(z, 0)$ & $[0.99, -0.24, 0.25, -0.69]$ \\
        $\Sigma_p(z, 0)$ & $[0.002, 0.009, 0.014, 0.025]$  \\
        $p(z, 0)$ & $[0.92, 0, 0.007, 0.073]$ \\
        \hline
    \end{tabular}
    \caption{Learned parameters of the 1D diffusion in a 4-well potential.}
    \label{tab:1D_4well}
\end{table}

\begin{table}
    \centering
    \begin{tabular}[b]{c|c}
        Parameter & Learned value \\
        \hline
        $\alpha_z$ & $\begin{pmatrix}
        1.05 & 0.07 \\ 0.05 & 0.98
        \end{pmatrix}$ , $\begin{pmatrix}
        1.12 & 0.12 \\ 0.26 & 0.99
        \end{pmatrix}$, $\begin{pmatrix}
        0.96 & 0.20 \\ -0.01 & 0.98
        \end{pmatrix}$ \\
        $\beta_z$ & $\begin{pmatrix}
        1.07\\0.30
        \end{pmatrix}$, $\begin{pmatrix}
        -0.77\\-0.15
        \end{pmatrix}$, $\begin{pmatrix}
        0.0\\1.04
        \end{pmatrix}$ \\
        $\Lambda$ & $\begin{pmatrix} -0.77 & 0.28 & 0.49 \\ 0.26 & -0.73 & 0.47 \\
        0.82 & 0.71 & -1.43\end{pmatrix}$ \\
        $\mu_p(z, 0)$ &$\begin{pmatrix}
        1.18\\0.11
        \end{pmatrix}$, $\begin{pmatrix}
        -0.32\\-0.05
        \end{pmatrix}$, $\begin{pmatrix}
        -0.41\\0.14
        \end{pmatrix}$\\
        $\Sigma_p(z, 0)$ & $\begin{pmatrix}
        0.264 & 0.058 \\ 0.058 & 0.269
        \end{pmatrix}$ , $\begin{pmatrix}
        0.018 & 0.002 \\ 0.002 & 0.015
        \end{pmatrix}$, $\begin{pmatrix}
        0.144 & 0.016 \\ 0.016 & 0.097
        \end{pmatrix}$  \\
        $p(z, 0)$ & $[0.019, 0.540, 441]$\\
        \hline
    \end{tabular}
    \caption{Learned parameters of the 2D diffusion in a 3-well potential.}
    \label{tab:2D_3well}
\end{table}

\subsection{Switching Ion Channel Data}
\label{sec:app_exp_channels}
The experimental data have been obtained using a voltage of 140$\,\mathrm{mV}$ and a 
sampling frequency of 5$\,\mathrm{kHz}$ over a measurement period of 1$\,\mathrm{s}$.
We fix the observation noise to $\sqrt{\Sigma_{\mathrm{obs}}} = \sigma_{\mathrm{obs}}= 0.25\,\mathrm{fA}$.
The full list of parameters is given in \cref{tab:ionchannels}. Initialization is 
done as in \cref{sec:app_exp_synSSDE}, but with $\Lambda_{z, z'} = 100$ for $z\neq z'$.
\begin{table}
    \centering
    \begin{tabular}[b]{c|c}
        Parameter & Learned value \\
        \hline
        $\alpha_z$ & $[1, 1, 1]$ \\
        $\beta_z$ & $[6.2 \cdot 10^{-12}, 4.4 \cdot 10^{-14}, 3.8 \cdot 10^{-12}]$\\
        $\Lambda$ & $\begin{pmatrix} -13.01 & 2.86 & 10.15\\  11.58 & -30.16 & 18.58\\
        44.33 & 23.35 & -67.68 \end{pmatrix}$\\
        $D$ & $[0.015, 0.001, 0.003, 0.01]$ \\
        $\mu_p(z, 0)$ & $[6.2 \cdot 10^{-12}, 4.4 \cdot 10^{-14}, 3.8 \cdot 10^{-12}]$ \\
        $\sqrt{\Sigma_p(z, 0)}$ & $[2.9 \cdot 10^{-13}, 3.7 \cdot 10^{-13}, 6.1 \cdot 10^{-13}]$  \\
        $p(z, 0)$ & $[1, 0, 0]$ \\
        \hline
    \end{tabular}
    \caption{Learned parameters for ion channel data.}
    \label{tab:ionchannels}
\end{table}

\subsection{Learning Complex Latent Continuous Dynamics}
\label{sec:app_exp_swirls}
Here, observations are drawn from a Poisson point process as in \cref{sec:app_exp_synSSDE}
with intensity $\lambda = 14$. The ground-truth and learned parameters are summarized in 
\cref{tab:swirls}. Initialization is done as in \cref{sec:app_exp_synSSDE}.
\begin{table}
    \centering
    \begin{tabular}[b]{c|c|c}
        Parameter & Ground truth value & Learned value \\
        \hline
        $\alpha_z$ & $\begin{pmatrix}
            0.6 & -1.4 \\ 2.6 & 0.6
            \end{pmatrix}$ , $\begin{pmatrix}
             -0.1 & 1.4 \\ -2.6 & 0.6
            \end{pmatrix}$ &
            $\begin{pmatrix}
            0.52 & -0.92 \\ 0.94 & -0.05
            \end{pmatrix}$ , $\begin{pmatrix}
            -0.34 & 1.41 \\ -1.98 & 0.37
            \end{pmatrix}$\\
        $\beta_z$ & $\begin{pmatrix}
            -5\\0
            \end{pmatrix}$, $\begin{pmatrix}
            5\\0
            \end{pmatrix}$ &
            $\begin{pmatrix}
            -6.08\\-1.69
            \end{pmatrix}$, $\begin{pmatrix}
            5.75\\-2.11
            \end{pmatrix}$\\
        $\Lambda$ & $\begin{pmatrix} -0.3 & 0.3  \\ 0.3, -0.3 \end{pmatrix}$ & $\begin{pmatrix} -0.17 & 0.17  \\ 0.23, -0.23 \end{pmatrix}$ \\
        $\mu_p(z, 0)$ &$\begin{pmatrix}
            -5\\0
            \end{pmatrix}$, $\begin{pmatrix}
            5\\0
            \end{pmatrix}$ & $\begin{pmatrix}
            -5.34\\-0.34
            \end{pmatrix}$, $\begin{pmatrix}
            5.61\\1.39
            \end{pmatrix}$\\
        $\Sigma_p(z, 0)$ & $\begin{pmatrix}
            0.49 & 0 \\ 0 & 0.49
            \end{pmatrix}$ , $\begin{pmatrix}
            0.49 & 0 \\ 0 & 0.49
            \end{pmatrix}$ &  $\begin{pmatrix}
            6.92 & 1.56 \\ 1.56 & 16.22
            \end{pmatrix}$ , $\begin{pmatrix}
            0.005 & 0.0002 \\ 0.0002 & 0.003
            \end{pmatrix}$  \\
        $p(z, 0)$ & $[0, 1]$ & $[0, 1]$\\
        \hline
    \end{tabular}
    \caption{Ground-truth and learned parameters of the complex structured 2D switching diffusion.}
    \label{tab:swirls}
\end{table}